\documentclass{article}

\usepackage[usenames,dvipsnames]{xcolor}

\usepackage{nameref,zref-xr}
\zxrsetup{toltxlabel}
\usepackage[colorlinks]{hyperref}       
\hypersetup{citecolor=MidnightBlue}
\hypersetup{linkcolor=MidnightBlue}
\hypersetup{filecolor=MidnightBlue}
\hypersetup{urlcolor=MidnightBlue}
\usepackage{url}            

\newcommand{\red}[1]{\textcolor{BrickRed}{#1}}
\newcommand{\orange}[1]{\textcolor{BurntOrange}{#1}}

\newcommand{\blue}[1]{\textcolor{MidnightBlue}{#1}}

\usepackage[disable]{todonotes}
\newcommand{\LT}[2][]{\todo[author=L.~Tiao,color={blue!100!green!33},size=\tiny,#1]{#2}}

\newcommand{\CA}[2][]{\todo[author=C.~Archambeau,color={gray!100!green!33},size=\tiny,#1]{#2}}

\usepackage[utf8]{inputenc} 

\usepackage{amsmath}
\usepackage{amssymb}
\usepackage{mathtools}
\usepackage{cancel}
\usepackage{empheq}

\newcommand{\mathbold}[1]{\ensuremath{\boldsymbol{\mathbf{#1}}}}

\usepackage[most]{tcolorbox}
\tcbset{highlight math style={colframe=Black,colback=Gray!5}}

\usepackage{microtype}
\usepackage{graphicx}
\usepackage{subcaption}
\usepackage{booktabs} 
\usepackage{multirow}

\usepackage{wrapfig}

\usepackage{siunitx}
\usepackage{nicefrac}       

\usepackage[algo2e,ruled,vlined,linesnumbered,commentsnumbered]{algorithm2e}

\SetCommentSty{mycommfont}

\SetNlSty{mynlfont}{}{} 
\SetSideCommentRight
\DontPrintSemicolon

\usepackage[acronym,smallcaps]{glossaries}
\glsdisablehyper

\DeclareRobustCommand{\parhead}[1]{\textbf{#1}~}

\usepackage[capitalise,noabbrev]{cleveref}
\crefname{step}{step}{steps}
\Crefname{step}{Step}{Steps}
\crefname{section}{\S}{\S\S}
\Crefname{section}{\S}{\S\S}
\crefformat{equation}{#2eq.~#1#3}
\Crefformat{equation}{#2Eq.~#1#3}

\newcommand{\g}{\,|\,}

\newcommand{\nestedmathbold}[1]{{\mathbold{#1}}}

\newcommand{\mbb}{\nestedmathbold{b}}

\newcommand{\mbf}{\nestedmathbold{f}}

\newcommand{\mbp}{\nestedmathbold{p}}

\newcommand{\mbx}{\nestedmathbold{x}}
\newcommand{\mby}{\nestedmathbold{y}}
\newcommand{\mbz}{\nestedmathbold{z}}

\newcommand{\mbW}{\nestedmathbold{W}}
\newcommand{\mbX}{\nestedmathbold{X}}

\newcommand{\mblambda}{\nestedmathbold{\lambda}}

\newcommand{\mbomega}{\nestedmathbold{\omega}}

\newcommand{\mbtheta}{\nestedmathbold{\theta}}

\DeclareMathOperator*{\argmax}{arg\,max}
\DeclareMathOperator*{\argmin}{arg\,min}

\newcommand{\cD}{\mathcal{D}}

\newcommand{\cL}{\mathcal{L}}
\newcommand{\cM}{\mathcal{M}}
\newcommand{\cN}{\mathcal{N}}
\newcommand{\cO}{\mathcal{O}}

\newcommand{\cX}{\mathcal{X}}

\newcommand{\bbE}{\mathbb{E}}
\newcommand{\bbI}{\mathbb{I}}

\newcommand{\bbR}{\mathbb{R}}

\newcommand{\defeq}{\coloneqq}

\newcommand{\Bernoulli}{\mathrm{Bernoulli}}

\newacronym[longplural={deep Gaussian processes}]{DGP}{dgp}{deep Gaussian process}
\newacronym[longplural={Gaussian processes}]{GP}{gp}{Gaussian process}
\newacronym{ABLR}{ablr}{adaptive Bayesian linear regression}
\newacronym{ADAM}{adam}{adaptive moment estimation}
\newacronym{ARD}{ard}{automatic relevance determination}
\newacronym{AUTOML}{automl}{automated machine learning}
\newacronym{BANANAS}{bananas}{Bayesian optimization with neural architectures for \textsc{nas}}
\newacronym{BCE}{bce}{binary cross-entropy}
\newacronym{BED}{bed}{Bayesian experimental design}
\newacronym{BNN}{bnn}{Bayesian neural network}
\newacronym{BOHAMIANN}{bohamiann}{foo bar baz}
\newacronym{BOREMLP}{bore-mlp}{\textsc{bore mlp}}
\newacronym{BORERF}{bore-rf}{\textsc{bore random forest}}
\newacronym{BORESVM}{bore-svm}{\textsc{bore svm}}
\newacronym{BOREXGB}{bore-xgb}{\textsc{bore xgboost}}
\newacronym{BORE}{bore}{Bayesian optimization by density-ratio estimation}
\newacronym{BO}{bo}{Bayesian optimization}
\newacronym{CI}{ci}{confidence interval}
\newacronym{CMAES}{cma-es}{covariance matrix adaptation evolution strategy}
\newacronym{CPE}{cpe}{class-probability estimation}
\newacronym{DAG}{dag}{directed acyclic graph}
\newacronym{DE}{de}{differential evolution}
\newacronym{DNGO}{dngo}{foo bar baz}
\newacronym{DRE}{dre}{density-ratio estimation}
\newacronym{ECDF}{ecdf}{empirical cdf}
\newacronym{EI}{ei}{expected improvement}
\newacronym{ERM}{erm}{empirical risk minimization}
\newacronym{ES}{es}{entropy search}
\newacronym{FCNET}{fcnet}{fully-connected neural network}
\newacronym{GAN}{gan}{generative adversarial network}
\newacronym{HB}{hb}{Hyperband}
\newacronym{KDE}{kde}{kernel density estimation}
\newacronym{KG}{kg}{knowledge gradient}
\newacronym{KLIEP}{kliep}{\textsc{kl} importance estimation procedure}
\newacronym{KMM}{kmm}{kernel mean matching}
\newacronym{KRR}{krr}{kernel ridge regression}
\newacronym{LBFGS}{l-bfgs}{limited-memory Broyden-Fletcher-Goldfarb-Shanno}
\newacronym{LOGREG}{logreg}{logistic regression}
\newacronym{LRT}{lrt}{likelihood-ratio testing}
\newacronym{LR}{lr}{learning rate}
\newacronym{LS}{ls}{local search}
\newacronym{LSTM}{lstm}{long short-term memory}
\newacronym{LSV}{lsv}{last-seen value}
\newacronym{MAP}{map}{maximum \textit{a posteriori}}
\newacronym{MCMC}{mcmc}{Markov chain Monte Carlo}
\newacronym{MC}{mc}{Monte Carlo}
\newacronym{MI}{mi}{mutual information}
\newacronym{MLE}{mle}{maximum likelihood estimation}
\newacronym{MLP}{mlp}{multi-layer perceptron}
\newacronym{MSE}{mse}{mean-squared error}
\newacronym{NAS}{nas}{neural architecture search}
\newacronym{NMSE}{nmse}{normalized mean-squared error}
\newacronym{NN}{nn}{neural network}
\newacronym{ODE}{ode}{ordinary differential equation}
\newacronym{OLS}{ols}{ordinary least squares}
\newacronym{PDE}{pde}{partial differential equation}
\newacronym{PES}{pes}{predictive \textsc{es}}
\newacronym{PI}{pi}{probability of improvement}
\newacronym{PROFET}{profet}{probabilistic data-efficient experimentation tool}
\newacronym{RE}{re}{regularized evolution}
\newacronym{RF}{rf}{random forest}
\newacronym{RL}{rl}{reinforcement learning}
\newacronym{RNN}{rnn}{recurrent neural network}
\newacronym{RS}{rs}{random search}
\newacronym{RULSIF}{rulsif}{relative \textsc{ulsif}}
\newacronym{SGD}{sgd}{stochastic gradient descent}
\newacronym{SGHMC}{sghmc}{stochastic gradient Hamiltonian Monte Carlo}
\newacronym{SH}{sh}{successive halving}
\newacronym{SMAC}{smac}{sequential model-based algorithm configuration}
\newacronym{SMBO}{smbo}{sequential model-based optimization}
\newacronym{SVM}{svm}{support vector machine}
\newacronym{SVR}{svr}{support vector regression}
\newacronym{TPE}{tpe}{tree-structured Parzen estimator}
\newacronym{UCB}{ucb}{upper confidence bound}
\newacronym{ULSIF}{ulsif}{unconstrained least-squares importance fitting}
\newacronym{XGBOOST}{xgboost}{gradient-boosted trees}

\usepackage[accepted,nohyperref]{icml2021}

\icmltitlerunning{Bayesian Optimization by Density-Ratio Estimation}

\begin{document}

\twocolumn[
\icmltitle{BORE: Bayesian Optimization by Density-Ratio Estimation}



\icmlsetsymbol{equal}{*}

\begin{icmlauthorlist}
\icmlauthor{Louis C.~Tiao}{usyd,d61}
\icmlauthor{Aaron~Klein}{amzn} 
\icmlauthor{Matthias~Seeger}{amzn} 
\icmlauthor{Edwin V.~Bonilla}{d61,usyd}
\icmlauthor{C\'{e}dric~Archambeau}{amzn}
\icmlauthor{Fabio~Ramos}{usyd,nvd}
\end{icmlauthorlist}

\icmlaffiliation{usyd}{University of Sydney, Sydney, Australia}
\icmlaffiliation{amzn}{Amazon, Berlin, Germany}
\icmlaffiliation{d61}{CSIRO's Data61, Sydney, Australia}
\icmlaffiliation{nvd}{NVIDIA, Seattle, WA, USA}

\icmlcorrespondingauthor{Louis~Tiao}{louis.tiao@sydney.edu.au}

\icmlkeywords{Machine Learning, ICML}

\vskip 0.3in
]



\printAffiliationsAndNotice{}  

\begin{abstract}
\gls{BO} is among the most effective and widely-used blackbox optimization 
methods. 
\gls{BO} proposes solutions according to an explore-exploit trade-off criterion 
encoded in an acquisition function,
many of which are 
computed from the 
posterior
predictive 
of a probabilistic surrogate model.
Prevalent among these is the \gls{EI}.
The need to ensure analytical tractability of the predictive often 
poses limitations that can 
hinder the efficiency and applicability of \gls{BO}.
In this paper,
we cast the computation of \gls{EI} as a binary classification problem, 
building on the link between \acrlong{CPE}
and \acrlong{DRE}, 
and the lesser-known link between density-ratios and \gls{EI}.
By 
circumventing the tractability constraints,
this reformulation provides numerous 
advantages, 
not least in 
terms of
expressiveness, 
versatility, and
scalability.
\CA{This needs to be more tangible. We need to quantify (give at least an example) the benefit!}
\LT{I need to emphasize that EI is usually expressed analytically as a combination of the properties of the posterior predictive, which is a major cause of the limitations--the need to ensure analytical tractability. And then I can explain that the increased flexibility and greater representational capacity comes from the alternate parameterization of EI as an arbitrarily powerful blackbox classifier, uninhibited by tractability constraints. Further, I need to say all this in one or two short sentences...}
\CA{We claim that the tractability constraints are circumvented. This is great, but raises the question why this imatters given that EI is not considered the SotA acquisition function. We need to make clear that EI is (is it?) now more competitive with other acquisition functions.}
\LT{If we claim that EI, formulated in the way we do, makes it more competitive against other acquisition functions, we need to run comparisons with a fairly large set of them---or at the very least, UCB, ES, and PES. Until we actually do this, I'd rather just say something cliche like: ``although EI is not the SOTA acquisition function, it is a practical, conceptually simple and effective choice that has withstood the test of time.''}
\CA{It would have been sufficient to run against one, let say PES or MES. We may actually do that after submission to prepare for the rebuttal :-)}
\end{abstract}

\glsresetall
\glsunset{BORE}
\glsunset{SMAC}

\begin{figure}[t]
  \vskip 0.2in
  \begin{center}
    \centerline{\includegraphics[width=\columnwidth]{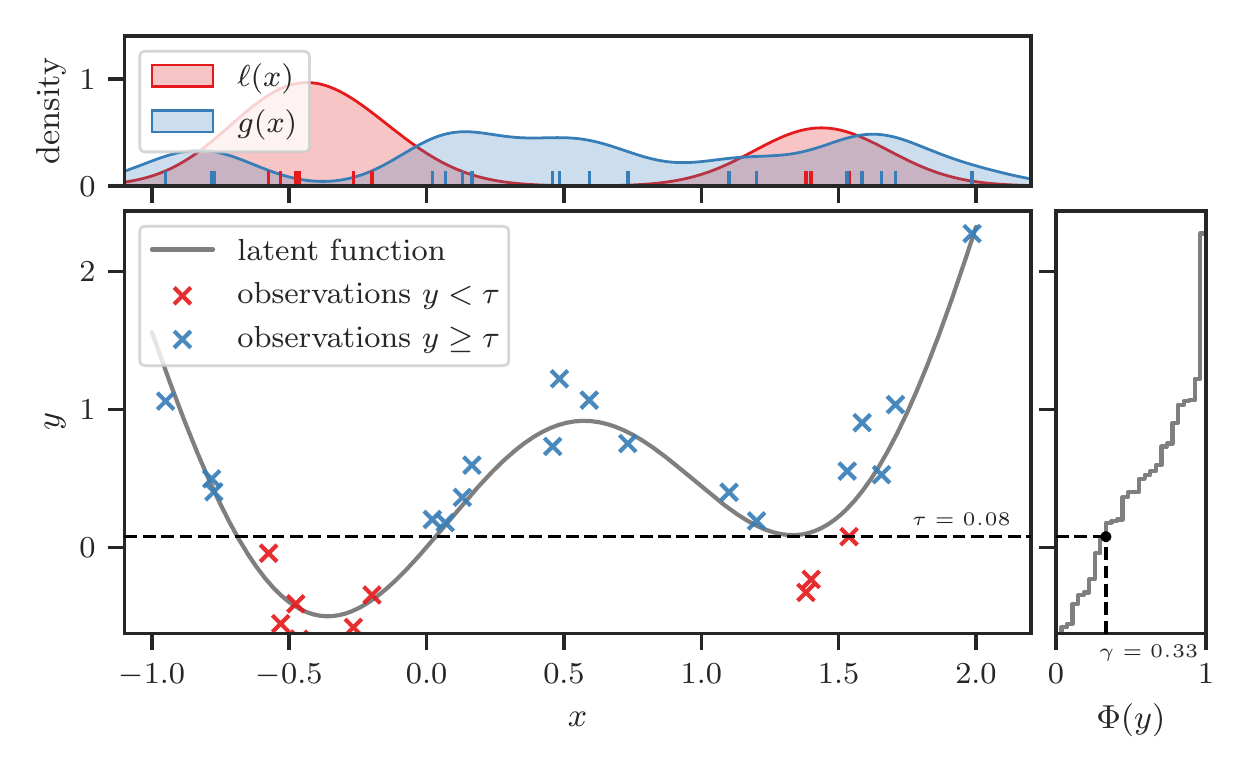}}
    \caption{Optimizing a synthetic function $f(x) = \sin(3x) + x^2 - 0.7x$
             with observation noise $\varepsilon \sim \cN(0, 0.2^2)$.
             In the main pane, the noise-free function is 
             represented by the solid gray curve, 
             and $N=27$ noisy observations are 
             represented by the crosses~`$\times$'.
             Observations with output $y$ in 
             the top performing $\gamma=\nicefrac{1}{3}$ proportion are shown in 
             \emph{red}; otherwise, they are shown in \emph{blue}. 
             Their corresponding 
             densities, 
             $\ell(x)$ and $g(x)$, 
             respectively, are shown in the top pane.
             \gls{BORE} exploits the correspondence between the \acrshort{EI} 
             acquisition function and the \emph{ratio} of densities 
             $\nicefrac{\ell(x)}{g(x)}$.}
             \CA{The last sentence is not informative and could confuse the reader.}
             \LT{True, but the reader need look no further than \cref{eq:ei-as-relative-density-ratio} to get the specifics; the teaser hopefully entices the reader to do that.}
    \label{fig:teaser}
  \end{center}
  \vskip -0.2in
\end{figure}

\section{Introduction}
\label{sec:introduction}

\gls{BO} is a 
sample-efficient
methodology for 
the optimization of 
expensive blackbox functions~\cite{brochu2010tutorial,shahriari2015taking}.
In brief, \gls{BO} proposes candidate solutions according to an 
\emph{acquisition function} that encodes the explore-exploit trade-off.
At the core of \gls{BO} is a probabilistic surrogate model
based on which the acquisition function can be computed.

Of the numerous acquisition functions that have been devised, 
the
\gls{EI}~\cite{mockus1978application,jones1998efficient} 
remains 
predominant,
due
in large 
to its effectiveness 
in spite of 
its relative simplicity.
In particular, while acquisition functions are
generally
difficult to 
compute, let alone optimize~\cite{wilson2018maximizing}, 
\gls{EI} 
has
a closed-form expression 
when the 
model's
posterior predictive 
is Gaussian.  
However, while this 
condition
makes \gls{EI} easier to 
work with, 
it can also
preclude 
the 
use
of richer families of 
models: 
one must 
ensure analytical 
tractability 
of the predictive,
often at the expense of expressiveness, 
or otherwise resort to sampling-based approximations. 

For instance,
by virtue of
its flexibility and
well-calibrated predictive uncertainty 
(not to mention conjugacy properties),
\gls{GP} regression~\cite{williams1996gaussian}
is a 
widely-used 
probabilistic model in \gls{BO}.
To extend 
\gls{GP}-based \gls{BO} to 
functions with
discrete variables~\cite{garrido2020dealing}, 
structures with conditional dependencies~\cite{jenatton2017bayesian}, 
or to capture
nonstationary phenomenon~\cite{snoek2014input},
it is common to 
apply simple 
modifications to
the covariance function, 
as this 
can often be done without
compromising the 
tractability 
of the 
predictive.
Suffice it to say, there exists estimators more naturally adept at 
dealing with these conditions (e.g. decision trees in the case of discrete variables).
Indeed,
to scale \gls{BO} to 
problem
settings
that 
yield large numbers of observations,
such as 
in
transfer learning~\cite{swersky2013multi}, 
existing works
have resorted to 
different families of models, 
such as \glspl{RF} \cite{hutter2011sequential} and
\glspl{BNN} \cite{snoek2015scalable,springenberg2016bayesian,perrone2018scalable}.
However, these are either subject to constraints and simplifying assumptions, or must 
resort to \gls{MC} 
methods
that make \gls{EI} more cumbersome to evaluate and optimize.

\LT{This is true and a good point, but I don't think we can claim that BORE does any better in this regard (it's a mystery how exploration actually factors into this density-ratio formulation of EI)
On the other hand, with BORE we can deploy models without worrying about making sure the posterior predictive is Gaussian. This is the point I really want to make with regard to NNs and RFs in BO.}
\CA{You might want to cite https://arxiv.org/abs/1706.04599}
\LT{This reference is more about calibration of deterministic neural network classifiers and less about predictive uncertainty of Bayesian neural networks(?)}
Recognizing that the surrogate model is only a means to an end---namely, of 
formulating 
an acquisition function, 
we turn the spotlight away from the model 
and toward the acquisition function itself.
To this end, we seek an alternative formulation of \gls{EI}, specifically,
one that potentially opens the door to more powerful 
estimators 
for which the
predictive
would otherwise be unwieldy or simply intractable to compute.
\CA{The wording is inaccurate. Even is the predictive posterior is analytically tractable, this does not mean the acquisition function is, nor that it is easy to evaluate. You might also have a predictive that is analytically tractable, but result in an acquisition function that can be easily evaluated. What you want to say is that we seek to express the acquisition function in an alternate form that does not impose constraints on the surrogate model}
\LT{What you say applies to acquisition functions _beyond_ PI, EI and UCB, right? I am guessing you have ES, PES and max-value ES in mind? I think I just need to refer specifically to EI, rather than making claims about acquisition functions in general for which counterexamples and caveats abound.}
In particular,
\citet{bergstra2011algorithms} demonstrate that, remarkably, the \gls{EI} 
function can be expressed as the \emph{relative} ratio between two densities 
\cite{yamada2011relative}. 
To estimate this ratio, they propose a method known as the \gls{TPE},
which naturally handles discrete and tree-structured inputs, and scales 
linearly with the number of observations.
In spite of its many advantages, however,
\gls{TPE} 
is not without deficiencies.

This paper makes the following contributions.
(i) In~\cref{sec:background} we 
revisit
the \gls{TPE} approach 
from first principles and identify its 
shortcomings for 
tackling 
the general \gls{DRE} problem.
(ii) In~\cref{sec:methodology} we propose a simple yet powerful alternative 
that casts the computation of \gls{EI} as probabilistic classification.
This approach is built on the 
aforementioned link between \gls{EI} and the relative density-ratio, and the
correspondence between \gls{DRE} and \gls{CPE}.
As such, it
retains the strengths of the \gls{TPE} method while ameliorating many of its weaknesses.
Perhaps most significantly, it
enables one
to leverage 
virtually any state-of-the-art classification method 
available.
\CA{I just commented out the above; too verbose and little added value.}
\LT{I will leave it out for now then, but I feel we do need to circle back to how this better addresses the limitations of GP-based BO: namely mixed continuous and discrete inputs, non-stationarities, heteroscedasticity and so on. Otherwise there is no point in spelling those out in the earlier paragraph.}
In ~\cref{sec:related_work} we discuss how our work relates to the 
existing state-of-the-art
methods for blackbox optimization and demonstrate,
through comprehensive experiments 
in~\cref{sec:experiments}, that our approach competes well with these 
methods on a diverse range of problems.
\CA{Again you need to be more concrete about the benefits/gains!}

\glsreset{EI}

\section{Background}
\label{sec:background}
\CA{You should restructure the Background section. My suggestion would be: 2.1 BO with EI; 2.2 Relation between EI and density ratio; 2.3 TPE.  The density ratio paragraph could be integrated into 2.2 and pitfalls in 2.3.}

Given a blackbox function $f : \cX \to \bbR$,
the goal of \gls{BO} is to find an input $\mbx \in \cX$ 
at which it is minimized,
given a set of 
$N$ 
input-output observations
$\cD_N = \{ (\mbx_n, y_n) \}_{n=1}^N$,
where output
$y_n = f(\mbx_n) + \varepsilon$ 
is 
assumed to be
observed with noise $\varepsilon \sim \cN(0, \sigma^2)$.
In particular,
having specified
a probabilistic surrogate model $\cM$, its posterior 
predictive~$p(y \g \mbx, \cD_N)$ is used to 
compute
the acquisition function $\alpha(\mbx; \cD_N)$,
a criterion that encapsulates the 
explore-exploit trade-off.
Accordingly,
candidate solutions
are obtained 
by maximizing this criterion,
$\mbx_{N+1} = \argmax_{\mbx \in \cX} \alpha(\mbx; \cD_N)$.
We now focus our discussion on the \gls{EI} function. 

\subsection{Expected improvement (EI)}
\label{sub:expected_improvement}

We first specify a utility function that quantifies the nonnegative 
amount by which 
~$y$ improves upon some threshold~$\tau$, $U(\mbx, y, \tau) \defeq \max(\tau - y, 0)$.
Then, the \gls{EI} function~\cite{mockus1978application} 
\CA{I would simplify and get rid of $\Phi$ if possible; you can always use $p(y<\tau)$. Why using $I$ for utility and not $U$? Most importantly, the notion of improvement has not been introduced; you need to explain why you call $ \max(\tau - y, 0)$ improvement.}
\LT{``The improvement is the amount by which $y$ decreases over $\tau$'' I feel
spelling out what improvement means is both more verbose and less precise than the exact expression $\tau - y$, which we already provide.}
is defined as the expected value of $U(\mbx, y, \tau)$ over 
the predictive
\begin{equation} \label{eq:expected-improvement-generic}
  \alpha(\mbx; \cD_N, \tau) \defeq \bbE_{p(y \g \mbx, \cD_N)}[U(\mbx, y, \tau)].
\end{equation}
By convention, 
$\tau$ is set to the \emph{incumbent}, or 
the lowest function value
so far observed 
$\tau=\min_n y_n$~\cite{wilson2018maximizing}.
Suppose the predictive takes the form of a Gaussian,
\begin{equation} \label{eq:posterior-predictive-gaussian}
  p(y \g \mbx, \cD_N) = \cN(y \g \mu(\mbx), \sigma^2(\mbx)).
\end{equation} 
This leads to
\begin{equation} \label{eq:expected-improvement-gaussian}
  \alpha(\mbx; \cD_N, \tau) = 
  \sigma(\mbx) \cdot \left [ \nu(\mbx) \cdot \Psi(\nu(\mbx)) + \psi(\nu(\mbx)) \right ],
\end{equation}
where $\nu(\mbx) \defeq \frac{\tau - \mu(\mbx)}{\sigma(\mbx)}$, and 
$\Psi, \psi$ denote the cdf and pdf of the 
normal distribution, respectively.
While this exact expression is both easy to evaluate and optimize, the 
conditions necessary to satisfy \cref{eq:posterior-predictive-gaussian} can 
often
come at the expense of flexibility and expressiveness.
Instead, let us consider a 
fundamentally
different way to express \gls{EI} itself.

\CA{In what sense is this a relaxation?}
\LT{Rather than defining the incumbent $\tau$ as the \emph{very best} value observed so far, it is ``relaxed'' to be the $\gamma$-th percentile of the observed values for some $\gamma > 0$.}
\CA{You need to say that the goal is to maximize EI. You might want to repeat the exploit explore trade-off and say that the conventional setting makes an implicit choice. Different choices can be made changing the threshold. You could also cite this paper by James Wilson, et al.: Maximizing acquisition functions for Bayesian optimization.}
\subsection{Relative density-ratio}
Let $\ell(\mbx)$ and $g(\mbx)$ be a pair of densities.
The \emph{$\gamma$-relative density-ratio} of $\ell(\mbx)$ and $g(\mbx)$ is 
defined as
\begin{equation} \label{eq:density-ratio-relative}
  r_{\gamma}(\mbx) \defeq \frac{\ell(\mbx)}{\gamma \ell(\mbx) + (1 - \gamma) g(\mbx)},
\end{equation}
where $\gamma \ell(\mbx) + (1 - \gamma) g(\mbx)$ denotes the 
\emph{$\gamma$-mixture density} with mixing proportion $0 \leq \gamma < 1$ 
\cite{yamada2011relative}.
Note that for $\gamma=0$, we recover the \emph{ordinary} density-ratio
$r_0(\mbx) = \nicefrac{\ell(\mbx)}{g(\mbx)}$.
Further, observe that 
$r_{\gamma}(\mbx) = h_{\gamma}(r_0(\mbx))$ where
$h_{\gamma} : u \mapsto \left ( \gamma + u^{-1} (1 - \gamma) \right )^{-1}$
for $u > 0$.
\CA{The use of $\gamma$ is confusion; why would this correspond to quantiles of $y$? Btw you might want to say that $u$ is non-negative.} 

We now discuss the conditions under which \gls{EI} 
can be expressed 
as the
ratio in \cref{eq:density-ratio-relative}.
First, set the threshold $\tau$ as the $\gamma$-th~quantile of the 
observed $y$ values,
$\tau \defeq \Phi^{-1}(\gamma)$ where 
$ \gamma = \Phi(\tau) \defeq p(y \leq \tau)$.
\LT{We've dropped the conditioning on $\cD_N$ here. Maybe mention this.}
Thereafter,
define the pair of
densities as
$\ell(\mbx) \defeq p \left (\mbx \g y \leq \tau; \cD_N \right )$ and 
$g(\mbx) \defeq p \left (\mbx \g y > \tau; \cD_N \right )$.
An illustrated example is shown in \cref{fig:teaser}.
Under these conditions, \citet{bergstra2011algorithms} demonstrate that
the \gls{EI} 
function can be expressed as the 
relative
density-ratio, up to some 
constant 
factor
\begin{equation} \label{eq:ei-as-relative-density-ratio}
  \alpha \left (\mbx; \cD_N , \Phi^{-1}(\gamma) \right ) \propto r_{\gamma}(\mbx).
\end{equation}
For completeness, we provide a self-contained derivation in \cref{sec:derivation}. 
Thus, this reduces the problem of maximizing \gls{EI} to that of maximizing the
relative
density-ratio, 
\begin{align} 
  \mbx_{N+1}
  & = \argmax_{\mbx \in \cX} \alpha \left (\mbx; \cD_N, \Phi^{-1}(\gamma) \right ) \nonumber \\
  & = \argmax_{\mbx \in \cX} r_{\gamma}(\mbx) \label{eq:maximize-relative-density-ratio}. 
\end{align}
To estimate the unknown 
relative
density-ratio,
one can appeal to a wide 
variety of approaches from the \gls{DRE} literature~\cite{sugiyama2012density}. 
We refer to this strategy as \acrfull{BORE}.

\glsreset{TPE}

\subsection{Tree-structured Parzen estimator}
\label{sub:tpe}

The \gls{TPE} \cite{bergstra2011algorithms} is an instance of the 
\gls{BORE} framework that seeks to solve the optimization problem of 
\cref{eq:maximize-relative-density-ratio} by taking the following approach:
\begin{enumerate}
  \item \label[step]{itm:ordinary}
    Since 
    $r_{\gamma}(\mbx) = h_{\gamma}(r_0(\mbx))$ where 
    $h_{\gamma}$ is strictly non-decreasing,
    focus instead on maximizing\footnote{$r_0(\mbx)$ denotes $\gamma=0$ solely in $r_{\gamma}(\mbx)$ of \cref{eq:density-ratio-relative}---it does \emph{not} signify threshold $\tau \defeq \Phi^{-1}(0)$, which would lead to density $\ell(\mbx)$ containing no mass. 
    We address this subtlety in \cref{sec:relative_density_ratio}.} $r_0(\mbx)$,
    \begin{equation*}
      \mbx_{\star} = \argmax_{\mbx \in \cX} r_0(\mbx).
    \end{equation*}
    \CA{Need to say non-decreasing as function of what.}
    \LT{This would become a little verbose; I think it is obvious from the context.}
    \item \label[step]{itm:kde} 
      Estimate the ordinary density-ratio $r_0(\mbx)$ by 
      separately estimating its constituent numerator $\ell(\mbx)$ and 
      denominator $g(\mbx)$, using a tree-based variant of 
      \gls{KDE}~\cite{silverman1986density}.
      \CA{Did you define KDE? Perhaps you could indicate what tree-based refers to?}
\end{enumerate}
It is not hard to see why \gls{TPE} might be favorable compared to 
methods based on \gls{GP} regression---one now incurs an $\cO(N)$ 
computational cost as opposed to the $\cO(N^3)$ cost of \gls{GP} posterior 
inference. 
Furthermore, it is 
equipped to deal with tree-structured, mixed 
continuous, ordered and unordered discrete inputs.
In spite of its 
advantages, \gls{TPE} is not without shortcomings. 

\subsection{Potential pitfalls}
\label{sub:potential_pitfalls}

\begin{figure}[t]
  \centering
  \begin{subfigure}[t]{0.49\columnwidth}
    \centering
    \includegraphics[width=\linewidth]{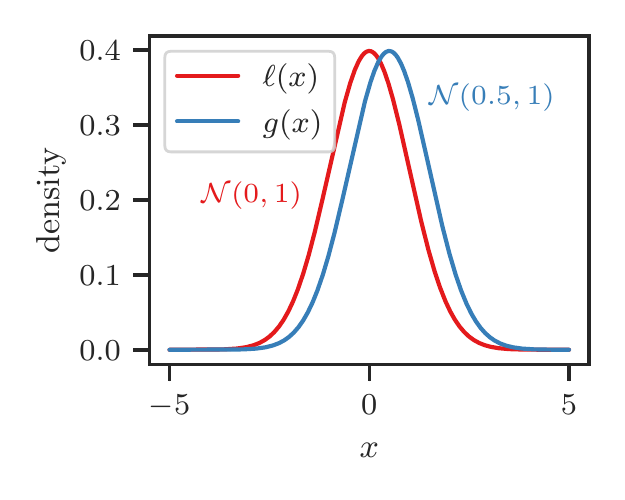}
    \caption{Densities}
  \end{subfigure}
  \begin{subfigure}[t]{0.49\columnwidth}
    \centering
    \includegraphics[width=\linewidth]{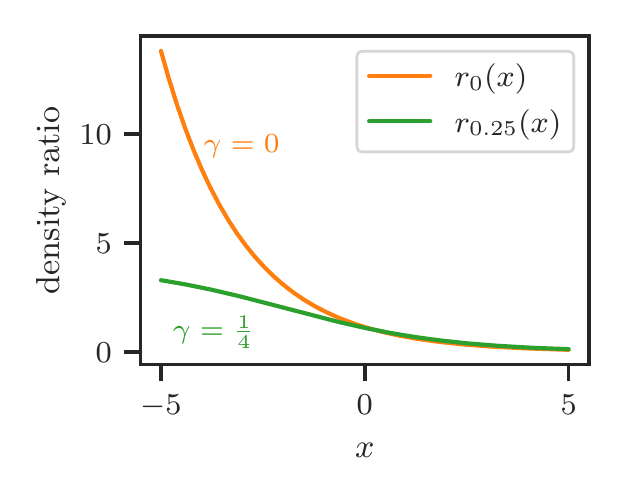}
    \caption{Density-ratios}
  \end{subfigure} 
  \caption{Gaussian densities 
  and their $\gamma$-relative density-ratios,
  which diverges when $\gamma=0$ and converges to 
  $4$ when $\gamma=\nicefrac{1}{4}$.}
  \label{fig:singularities}
\end{figure}
The shortcomings of this approach are already well-documented in the \gls{DRE} 
literature \cite{sugiyama2012density}. 
Nonetheless, we reiterate here a select few that are particularly detrimental 
in the context of global optimization. 
Namely, the first major drawback of \gls{TPE} lies within \cref{itm:ordinary}:

{\it Singularities.}
Relying on the ordinary density-ratio can result in numerical instabilities 
since it is unbounded---often diverging to infinity, even in simple toy scenarios (see \cref{fig:singularities} for a simple example).
In contrast, the $\gamma$-relative density-ratio is always bounded above by 
$\gamma^{-1}$ when $\gamma > 0$~\cite{yamada2011relative}.
The other potential problems of \gls{TPE} lie within \cref{itm:kde}:

{\it Vapnik's principle.}
Conceptually, independently estimating the densities is actually a more 
cumbersome approach that violates Vapnik's principle---namely, that when 
solving a problem of interest, one should refrain from resorting to solve a 
more general problem as an intermediate step~\cite{vapnik2013nature}. 
In this instance, \emph{density} estimation is a more general problem that is 
arguably more difficult than \emph{density-ratio} estimation \cite{kanamori2010theoretical}.

{\it Kernel bandwidth.} 
\gls{KDE} depends crucially on the selection of an appropriate kernel bandwidth,
which is notoriously difficult \cite{park1990comparison,sheather1991reliable}.
Furthermore, even with an optimal selection of a 
single
fixed 
bandwidth, it cannot simultaneously adapt to low- and high-density regions 
\cite{terrell1992variable}.

{\it Error sensitivity.} %
These difficulties are exacerbated by the fact that one is required to select
\emph{two} bandwidths, whereby the optimal bandwidth for one individual density 
is not necessarily appropriate for estimating the density-\emph{ratio}---indeed, 
it may even have deleterious effects.
This also makes the approach unforgiving to misspecification of the respective 
estimators, particularly in that of the denominator $g(\mbx)$, which has 
a disproportionately large 
influence on the resulting density-ratio. 

{\it Curse of dimensionality.}
For these reasons and more,
\gls{KDE} often falls short in high-dimensional regimes. 
In contrast, direct \gls{DRE} methods have consistently been shown to scale 
better
with dimensionality \cite{sugiyama2008direct}.

{\it Optimization.}
Ultimately, we care not only about \emph{estimating} the density-ratio, but 
also \emph{optimizing} it wrt to inputs for the purpose of candidate suggestion.
Being 
nondifferentiable, the ratio of \glspl{TPE} is cumbersome to optimize. 

\glsreset{CPE}
\glsreset{BCE}

\section{Methodology}
\label{sec:methodology}

We propose different approach to \gls{BORE}---importantly, one that 
circumvents the issues of \gls{TPE}---by 
seeking to \emph{directly} estimate the unknown ratio $r_\gamma(\mbx)$.
There exists a multitude of direct \gls{DRE} methods 
(see \cref{sec:related_work}).
Here, we focus on a conceptually simple and widely-used method based on \gls{CPE}
\cite{qin1998inferences,cheng2004semiparametric,bickel2007discriminative,sugiyama2012density,menon2016linking}.

First, let $\pi(\mbx) = p(z = 1 \g \mbx)$ denote the \emph{class-posterior probability}, 
where $z$ is the
binary class label
\begin{equation*}
  z \defeq
  \begin{cases} 
    1 & \text{if } y \leq \tau, \\
    0 & \text{if } y > \tau.
  \end{cases}
\end{equation*}
By definition, we have $\ell(\mbx) = p(\mbx \g z = 1)$ and $g(\mbx) = p(\mbx \g z = 0)$. 
\LT{Note we've dropped the conditioning on $\cD_N$ here.}
We plug these into \cref{eq:density-ratio-relative} and apply Bayes' rule, 
letting the $p(\mbx)$ terms cancel 
each other out 
to give
\begin{equation} \label{eq:relative-density-ratio-bayes-rule}
  \begin{split}
    & r_{\gamma}(\mbx) = \left ( \frac{p(z = 1 \g \mbx)}{p(z = 1)} \right ) \\
    & \times \left ( \gamma \cdot \frac{p(z = 1 \g \mbx)}{p(z = 1)} +
    (1 - \gamma) \cdot \frac{p(z = 0 \g \mbx)}{p(z = 0)} \right )^{-1}
  \end{split}
\end{equation}

Since $p(z = 1) = \gamma$ by definition, 
\cref{eq:relative-density-ratio-bayes-rule} 
simplifies 
to
\begin{equation} \label{eq:relative-density-ratio-class-posterior}
  r_{\gamma}(\mbx) 
  = \gamma^{-1} \pi(\mbx).
\end{equation}
Refer to \cref{sec:class_posterior_probability} for derivations.
Thus, \cref{eq:relative-density-ratio-class-posterior} establishes the link 
between the class-posterior probability and the relative density-ratio.
In particular, the 
latter
is equivalent to the 
former
up to constant factor $\gamma^{-1}$.

Let us estimate the 
probability $\pi(\mbx)$ using a 
probabilistic classifier---a function 
$\pi_\mbtheta : \cX \to [0, 1]$ 
parameterized by $\mbtheta$. 
To recover the true class-posterior probability, we minimize a 
\emph{proper scoring rule} \cite{gneiting2007strictly}, such as the log loss
\begin{equation} \label{eq:log-loss-empirical}
  \begin{split}
  \cL(\mbtheta) 
  & \defeq - \frac{1}{N} 
  \left ( \sum_{n=1}^N z_n \log{\pi_\mbtheta(\mbx_n)} \right . \\
  & \qquad \left . \vphantom{\sum_{n=1}^N} + (1 - z_n) \log{(1-\pi_\mbtheta(\mbx_n))} \right ).
  \end{split}
\end{equation}
Thereafter, we approximate the relative density-ratio up to constant $\gamma$ 
through
\begin{equation} \label{eq:approximation}
  \pi_\mbtheta(\mbx) \simeq \gamma \cdot r_{\gamma}(\mbx),
\end{equation}
with equality 
at $\mbtheta_{\star} = \argmin_{\mbtheta} \cL(\mbtheta)$. 
Refer to \cref{sec:log_loss} for derivations.
Hence, in the so-called \gls{BO} loop (summarized in \cref{alg:bo-loop}), we
alternately optimize (i) the classifier parameters $\mbtheta$ wrt to the log 
loss (to improve the approximation of \cref{eq:approximation}; \cref{line:parameters}), 
and (ii) the classifier input $\mbx$ wrt to its output (to suggest the next candidate to evaluate; \cref{line:inputs}).
An 
animation of \cref{alg:bo-loop} is provided in 
\cref{sec:step_through_visualization}. \CA{We should explain this in more length, referring to the relevant equations and repeating that step one corresponds to updating the posterior and step two to maximize EI.}
\LT{I do this in the next paragraph}
\CA{Btw it is not clear to me how you compute the expectations wrt to l(x) and g(x). We should say a few words on that too.}

In traditional \gls{GP}-based \gls{EI}, \cref{line:inputs} typically consists 
of maximizing the \gls{EI} function expressed in the form of \cref{eq:expected-improvement-gaussian},
while \cref{line:parameters} consists of optimizing the \gls{GP} 
hyperparameters wrt the marginal likelihood.
\LT{People might nitpick here and point out that it is also possible to sample from approximate posterior and other things that are besides the point...}
By analogy with our approach, the parameterized function $\pi_\mbtheta(\mbx)$
is \emph{itself} an approximation to the \gls{EI} function to be maximized 
directly, while the approximation is tightened through 
by optimizing the classifier parameters wrt the log loss.

In short, we have reduced the problem of computing \gls{EI} to that of 
learning a probabilistic classifier,
thereby unlocking a
broad 
range of estimators beyond those so far used in \gls{BO}.
Importantly, this enables one to employ virtually any state-of-the-art classification 
method available, and to parameterize the classifier using arbitrarily expressive 
approximators that 
potentially have the capacity to deal with 
non-linear, non-stationary, and 
heteroscedastic phenomena
frequently encountered in practice. 
\LT{This sentence is too long}

\begin{algorithm2e}[tb]
  \SetAlgoLined
\KwIn{blackbox $f : \cX \to \bbR$, 
      proportion $\gamma \in (0, 1)$, 
      probabilistic classifier $\pi_{\mbtheta} : \cX \to [0, 1]$.}
\While{under budget}{
  $\tau \gets \Phi^{-1}(\gamma)$ \tcp*{compute $\gamma$-th quantile of $\{ y_n \}_{n=1}^N$}
  $z_n \gets \bbI[y_n \leq \tau]$ for $n=1, \dotsc, N$ \tcp*{assign labels}
  $\tilde{\cD}_N \gets \{ (\mbx_n, z_n) \}_{n=1}^N$ \tcp*{construct auxiliary dataset}
  \tcc{update classifier by optimizing parameters $\mbtheta$ wrt log loss}
  $\mbtheta_{\star} \gets \argmin_\mbtheta \cL(\mbtheta)$ \tcp*{depends on $\tilde{\cD}_N$, see \cref{eq:log-loss-empirical}} \label{line:parameters} 
  \tcc{suggest candidate by optimizing input $\mbx$ wrt classifier} 
  $\mbx_{N} \gets \argmax_{\mbx \in \cX} \pi_{\mbtheta_{\star}}(\mbx)$ \tcp*{see \cref{eq:approximation}} \label{line:inputs}
  $y_{N} \gets f(\mbx_{N})$ \tcp*{evaluate blackbox function}
  $\cD_{N} \gets \cD_{N-1} \cup \{ (\mbx_{N}, y_{N}) \}$ \tcp*{update dataset}
  $N \gets N+1$
}
  \caption{\acrfull{BORE}.} 
  \label{alg:bo-loop}
\end{algorithm2e}



\CA{One might wonder why we still call Bayesian optimization.}
\LT{Agreed---just as people refer to TPE as a BO method, which is the apparently the subject of much controversy.}

\subsection{Choice of proportion $\gamma$}
\label{sub:choice_of_proportion}


The proportion $\gamma \in (0, 1)$ influences the explore-exploit trade-off.
Intuitively, a smaller setting of $\gamma$ encourages exploitation and leads to 
fewer modes and sharper peaks in the acquisition function.
To see this, consider that there are by definition fewer candidate inputs 
$\mbx$ for which its corresponding output $y$ can be expected to improve over 
the first quartile ($\gamma=\nicefrac{1}{4}$) of the observed output values 
than, say, the third quartile ($\gamma=\nicefrac{3}{4}$).
That being said, given that the class balance rate is by definition $\gamma$, 
a value too close to $0$ may lead to instabilities in classifier learning. 
A potential strategy to 
combat
this is to begin with a perfect balance ($\gamma=\nicefrac{1}{2}$) 
and then to decay $\gamma$ as optimization progresses.
In this work, we keep $\gamma$ fixed throughout optimization.
This, on the other hand, has the 
benefit of providing guarantees about how
the classification task evolves over 
iterations. 
In particular, in each iteration, after having observed a new evaluation, we 
are guaranteed that the binary label of \emph{at most} one existing 
instance can flip.
This property exploited to make classifier learning of 
\cref{line:parameters} more efficient by 
adopting
online learning 
techniques 
that avoid learning from scratch in each iteration.
An extended discussion is included in \cref{sec:properties_of_the_classification_problem}.

\subsection{Choice of probabilistic classifier} 
\label{sub:choice_of_probabilistic_classifier}




\LT{Note sure if this best fits here, or in the next (experiments) section.}

We examine a few 
variations of \gls{BORE} that differ in the choice of classifier and discuss 
their 
strengths and weaknesses across different 
global optimization problem settings.

\parhead{Multi-layer perceptrons.}
We propose \acrshort{BOREMLP}, a variant based on \glspl{MLP}.
This choice is appealing not only for 
(i) its flexibility and universal 
approximation guarantees \cite{hornik1989multilayer} but because
(ii) one can easily adopt \gls{SGD} methods to scale up its parameter learning 
\cite{lecun2012efficient}, and 
(iii) it is differentiable end-to-end, thus enabling the use of quasi-Newton 
methods such as \acrshort{LBFGS} \cite{liu1989limited} for candidate 
suggestion.
Lastly, 
since \gls{SGD} is online by nature,
(iv) it is feasible to adapt weights from previous iterations instead 
of training from scratch. 
A notable weakness is that \glspl{MLP} can be over-parameterized and 
therefore considerably 
data-hungry.
\LT{Another downside is that they are so-so for discrete spaces and conditional 
spaces (requires a lot more work, but can be done).}

\glsreset{RF}

\parhead{Tree-based ensembles.}
We propose two further variants: \acrshort{BORERF} and \acrshort{BOREXGB}, 
both based on ensembles of 
decision trees---namely, \gls{RF}~\cite{breiman2001random} 
and \gls{XGBOOST}~\cite{chen2016xgboost}, respectively.
These variants are attractive since they inherit from decision trees the 
ability to
(i) deal with discrete and conditional inputs by design,
(ii) work well in high-dimensions, and 
(iii) are scalable and easily parallelizable.
Further, (iv) online extensions of \glspl{RF}~\cite{saffari2009line} may be 
applied to avoid training from scratch.

A caveat is that, since their response surfaces are discontinuous and 
nondifferentiable, decision trees are difficult to maximize.
Therefore, we 
appeal to
random search and evolutionary strategies 
for
candidate suggestion.
Further details and a comparison of various approaches is included in \cref{sub:maximizing_the_acquisition_function}.

In theory, for the approximation of \cref{eq:approximation} to be tight, the 
classifier is required to produce well-calibrated probabilities~\cite{menon2016linking}.
A potential drawback of the \acrshort{BORERF} variant is that \glspl{RF} are 
generally
not trained by minimizing a proper scoring rule.
As such, 
additional techniques 
may be necessary to improve calibration~\cite{niculescu2005predicting}.



\CA{There should be a discussion about the class unbalance as well as the choice of $\gamma$.}

\begin{figure*}[ht]
  \centering
  \begin{subfigure}[t]{0.49\textwidth}
    \centering
    \includegraphics[width=\linewidth]{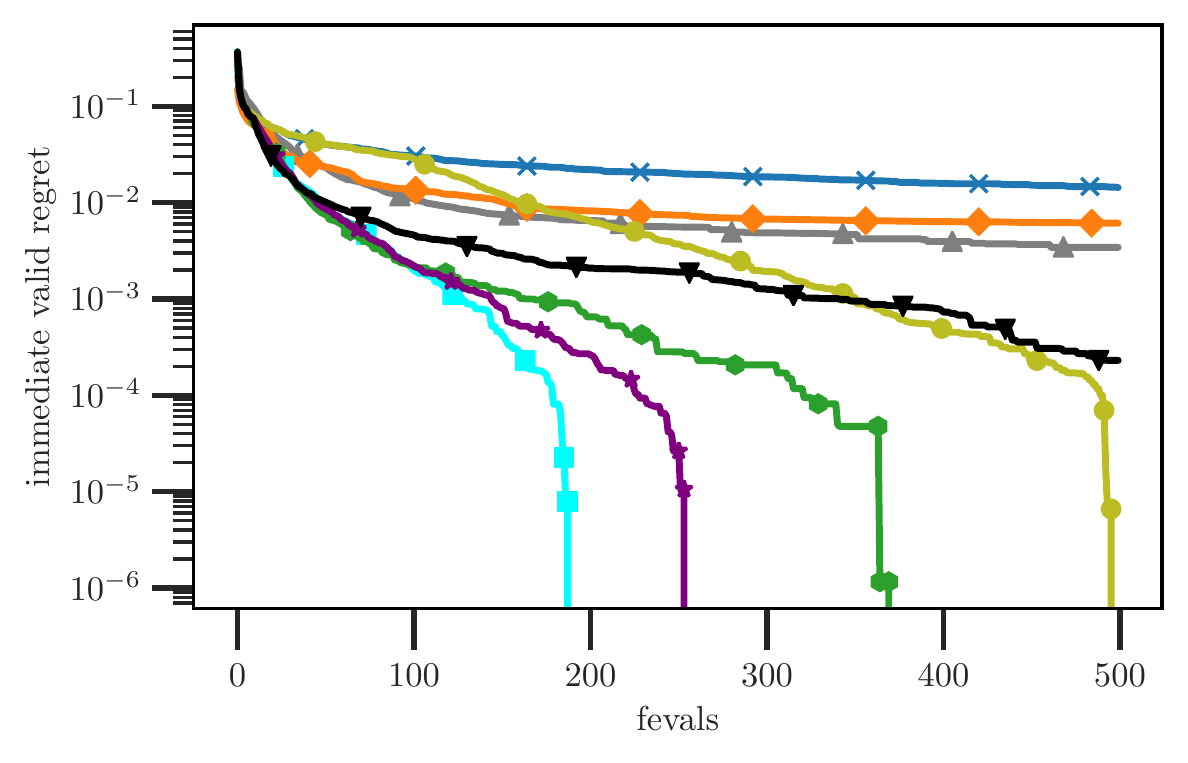}
    \caption{\textsc{protein}}
  \end{subfigure}
  ~
  \begin{subfigure}[t]{0.49\textwidth}
    \centering
    \includegraphics[width=\linewidth]{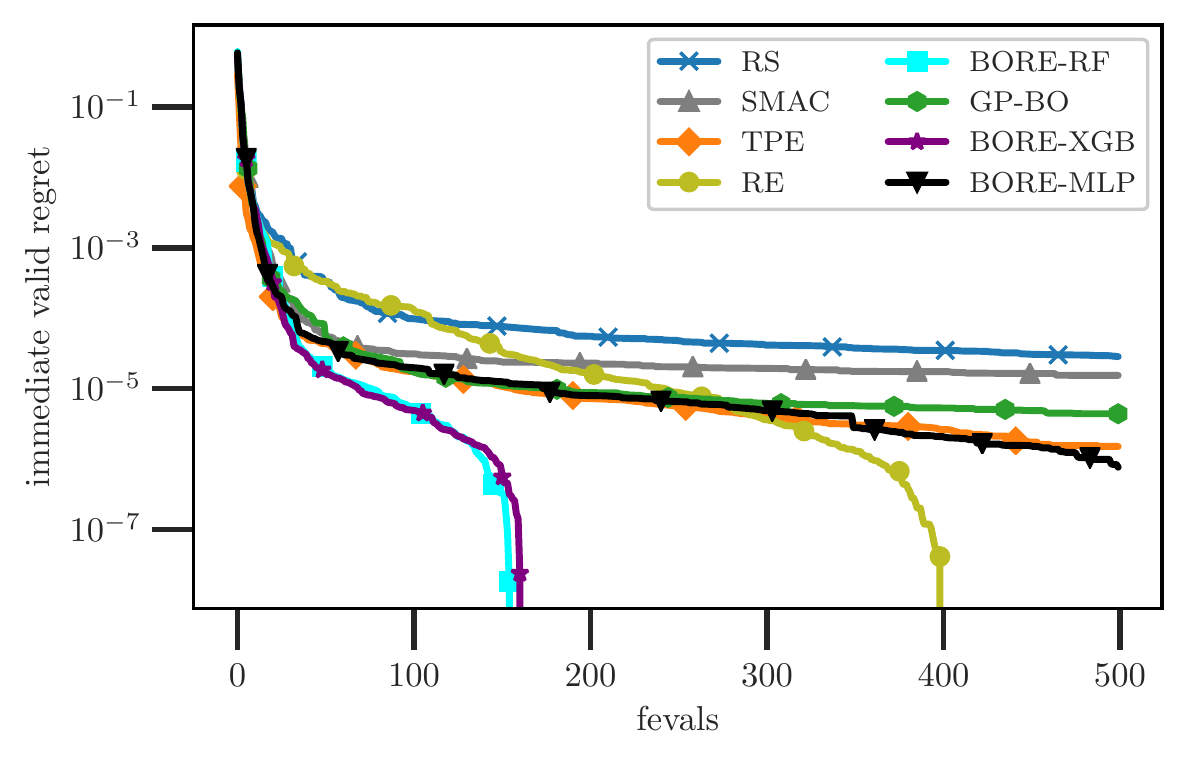}
    \caption{\textsc{naval}}
  \end{subfigure}
  \begin{subfigure}[t]{0.49\textwidth}
    \centering
    \includegraphics[width=\linewidth]{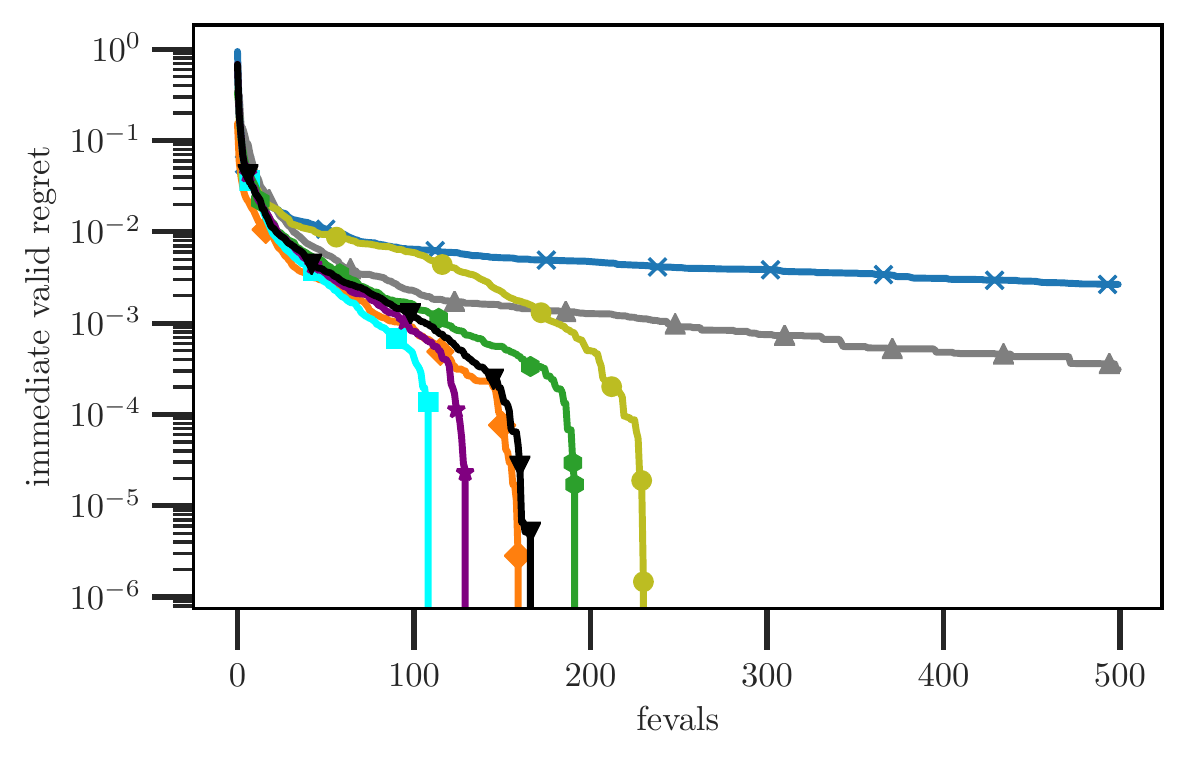}
    \caption{\textsc{parkinsons}}
  \end{subfigure}
  ~
  \begin{subfigure}[t]{0.49\textwidth}
    \centering
    \includegraphics[width=\linewidth]{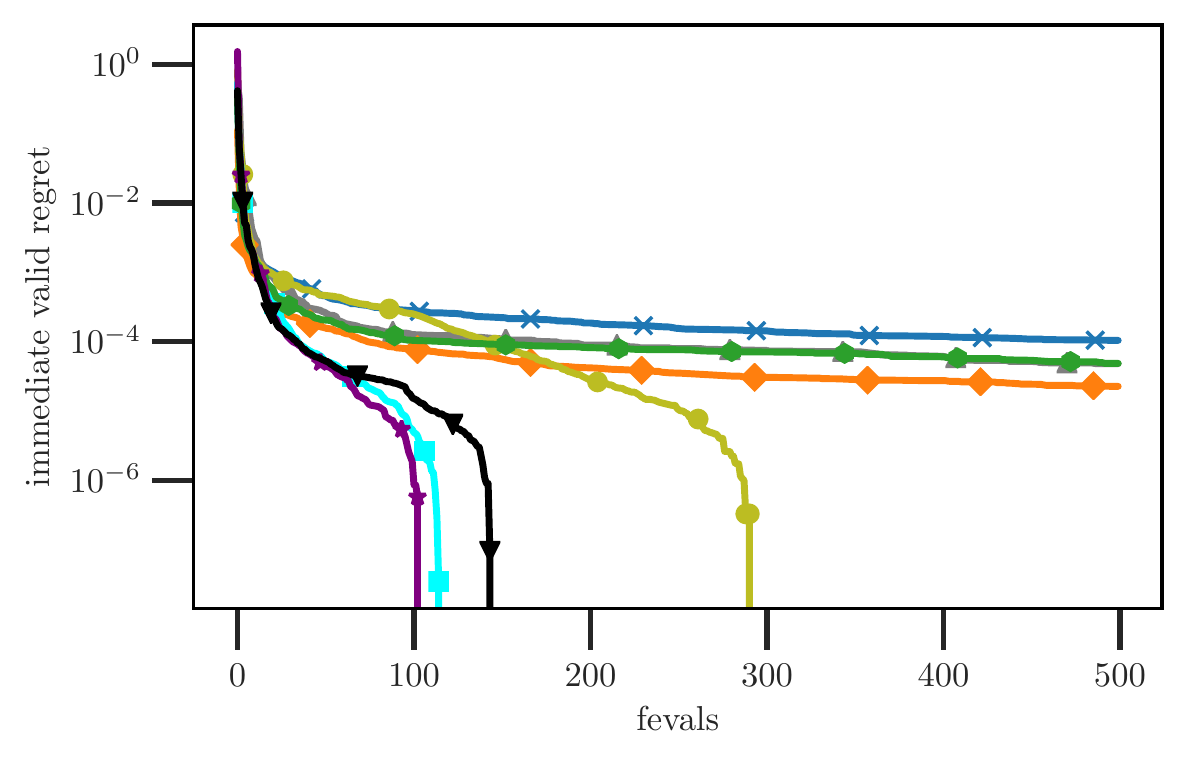}
    \caption{\textsc{slice}}
  \end{subfigure}
  \caption{Immediate regret over function evaluations on the HPOBench neural network tuning problems ($D=9$).}
  \label{fig:hpobench}
\end{figure*}

\section{Related Work}
\label{sec:related_work}

The literature on \gls{BO} is vast and ever-expanding \cite{brochu2010tutorial,shahriari2015taking,frazier2018tutorial}.
Some specific threads pertinent to our work include achieving scalability 
through \glspl{NN}, as in 
\acrshort{BANANAS}~\cite{white2019bananas},
\acrshort{ABLR}~\cite{perrone2018scalable},
\acrshort{BOHAMIANN}~\cite{springenberg2016bayesian}, and
\acrshort{DNGO}~\cite{snoek2015scalable},
and handling discrete and conditional variables using tree ensembles, 
as with \glspl{RF} in \glspl{SMAC}~\cite{hutter2011sequential}. 
To negotiate the tractability of the predictive, these methods 
must either make simplifications or resort to approximations.
In contrast, 
by seeking
to directly approximate the 
acquisition function, \gls{BORE} is unencumbered by such constraints.
Refer to \cref{sub:parameters} for an expanded discussion.
\LT{\cite{garrido2020dealing,snoek2014input} make patches to the kernel whereas 
we are free to adopt the class of estimators most appropriate to the problem
at hand. \red{mention this if we include in our comparisons}}
Beyond the 
classical
\acrshort{PI}~\cite{kushner1964new} and 
\gls{EI} functions~\cite{jones1998efficient},
a multitude of acquisition functions have been devised, including the
\gls{UCB}~\cite{srinivas2009gaussian},
\gls{KG}~\cite{scott2011correlated},
\gls{ES}~\cite{hennig2012entropy}, and 
\gls{PES}~\cite{hernandez2014predictive}.
\LT{The purpose of this paragraph is to (a) acknowledge that acquisition 
functions other than EI exist in BO, and (b) provide a justification for why 
one might still care to focus on EI with the existence of other, ostensibly better, options.}
Nonetheless, \gls{EI} remains ubiquitous, in large because it is conceptually 
simple, easy to evaluate 
and optimize, 
and consistently performs well in practice.


There is a substantial body of existing works on density-ratio estimation~\cite{sugiyama2012density}.
Recognizing the deficiencies of the \gls{KDE} approach, a 
myriad
alternatives have since been 
proposed, including
\gls{KLIEP}~\cite{sugiyama2008direct}, 
\gls{KMM}~\cite{gretton2009covariate},
\gls{ULSIF}~\cite{kanamori2009least}, and
\gls{RULSIF}~\cite{yamada2011relative}.
In this work, we restrict our focus on \gls{CPE}, an effective and versatile 
approach that has found widespread adoption in a diverse
range of 
applications, e.g. in 
covariate shift adaptation~\cite{bickel2007discriminative}, 
energy-based modelling~\cite{gutmann2012noise}, 
\glspl{GAN}~\cite{goodfellow2014generative,nowozin2016f}, 
likelihood-free inference~\cite{tran2017hierarchical,thomas2020likelihood}, and more.
Of particular relevance is its use in 
\gls{BED}, a close relative of \gls{BO}, wherein it is similarly applied to 
approximate the
expected utility function~\cite{kleinegesse2019efficient}.
\LT{I use up all this space to enumerate these other applications to justify the reason we adopt the CPE and not any of the others aforementioned. But perhaps this is a bit much.}

\section{Experiments}
\label{sec:experiments}

\glsreset{AUTOML}

We describe the experiments conducted to empirically evaluate our method.
To this end, we consider a variety of problems, ranging from 
\gls{AUTOML}, 
robotic arm
control, 
to racing line optimization. 


We provide comparisons against a comprehensive selection of state-of-the-art 
baselines. 
Namely, across all problems, we consider:
\gls{RS}~\cite{bergstra2012random},
\textsc{gp-bo} with \gls{EI}, $\gamma=0$~\cite{jones1998efficient},
\gls{TPE}~\cite{bergstra2011algorithms}, and
\gls{SMAC}~\cite{hutter2011sequential}.
We also consider evolutionary strategies: 
\gls{DE}~\cite{storn-jgo97} for problems with continuous domains, and
\gls{RE}~\cite{real2019regularized} for those with discrete domains.
Further information about these baselines and the source code for their 
implementations are included in \cref{sec:baselines_implementations}.

\newpage 
To quantitatively assess performance (on benchmarks for which the exact global 
minimum is known),
we report the \emph{immediate regret}, 
defined as the absolute error between the global minimum and the lowest 
function value attained thus far.
Unless otherwise stated we report, for each benchmark and method, results
aggregated
across 
100 
replicated 
runs. 

We set $\gamma=\nicefrac{1}{3}$ across all variants and
benchmarks. 
For candidate suggestion in the tree-based variants, we use \gls{RS} with a 
function evaluation limit of 500 for problems with discrete domains, and \gls{DE} 
with a limit of 2,000 for those with continuous domains.
Further details concerning the experimental set-up and the implementation of 
each \gls{BORE} variant are included in \cref{sec:details_implementations}.

\parhead{Neural network tuning (HPOBench).} 
First we consider the problem of training a two-layer 
feed-forward \gls{NN}
for regression.
Specifically, a \gls{NN} is trained for 100 epochs with the \acrshort{ADAM} 
optimizer \cite{kingma2014adam}, and the objective is the validation \gls{MSE}.
The hyperparameters are the \emph{initial learning rate}, 
\emph{learning rate schedule}, \emph{batch size}, along with the 
layer-specific \emph{widths}, \emph{activations} and \emph{dropout rates}.
\LT{Should emphasize that all inputs are discrete ordered and unordered}
We consider four 
datasets: 
\textsc{protein}, \textsc{naval}, \textsc{parkinsons} and \textsc{slice},
and utilize HPOBench~\cite{klein2019tabular} which
tabulates, 
for each 
dataset,
the 
\glspl{MSE} resulting from
all possible (62,208) configurations.
Additional details are included in \cref{sub:hpobench},
and the results are shown in \cref{fig:hpobench}. 
We see across all datasets that the \gls{BORE}-\textsc{rf} and -\textsc{xgb} 
variants consistently 
outperform
all other 
baselines,  
converging rapidly toward the global minimum after 1-2 hundred evaluations---in 
some cases, earlier
than any other 
baseline by 
over
two hundred evaluations.
Notably, with the exception being \acrshort{BOREMLP} on the \textsc{parkinsons} dataset, 
all \gls{BORE} variants outperform \gls{TPE}, in many cases by a sizable margin.

\begin{figure*}[ht]
  \centering
  \begin{subfigure}[t]{0.32\textwidth}
    \centering
    \includegraphics[width=\linewidth]{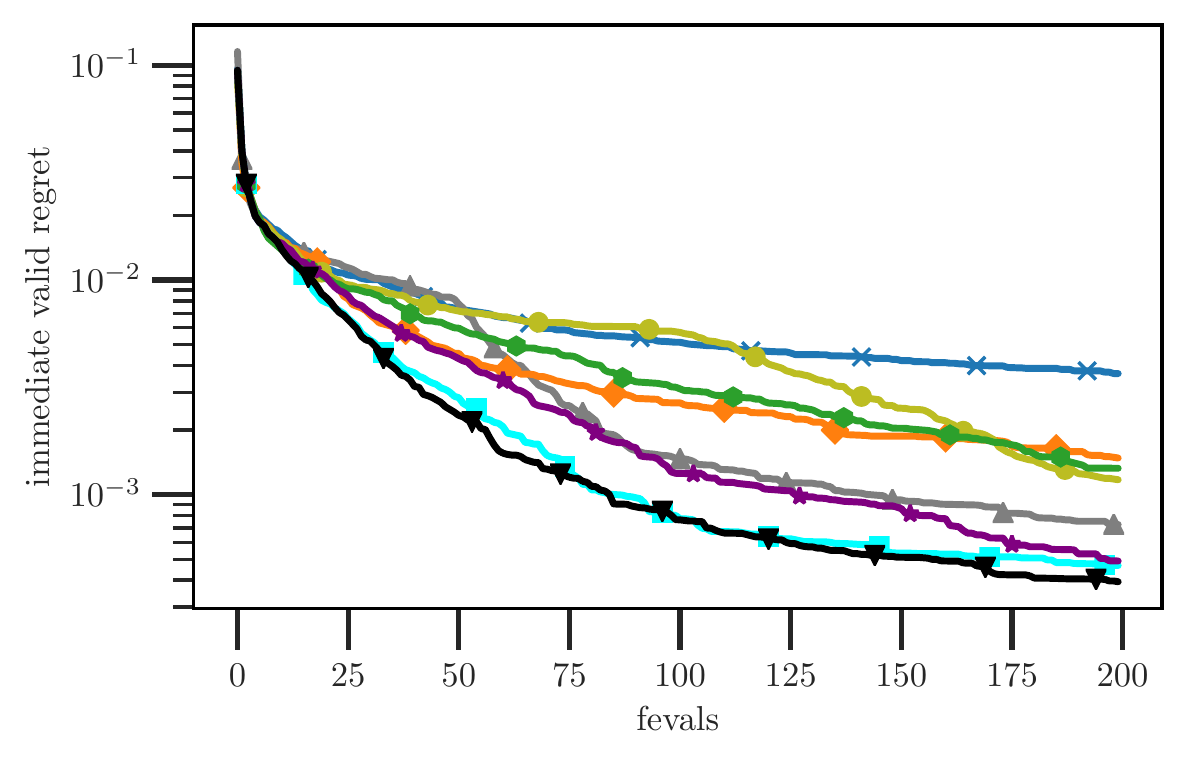}
    \caption{CIFAR-10}
  \end{subfigure}
  ~
   \begin{subfigure}[t]{0.32\textwidth}
    \centering
    \includegraphics[width=\linewidth]{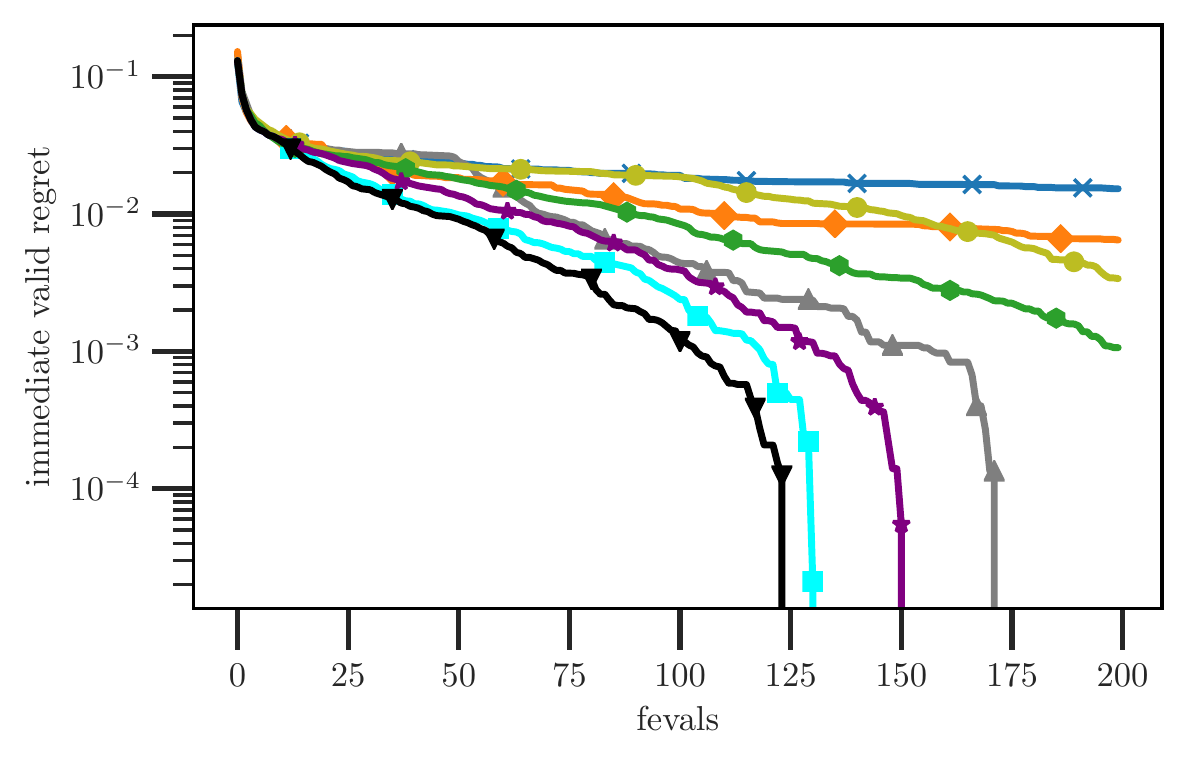}
    \caption{CIFAR-100}
  \end{subfigure} 
  ~
   \begin{subfigure}[t]{0.32\textwidth}
    \centering
    \includegraphics[width=\linewidth]{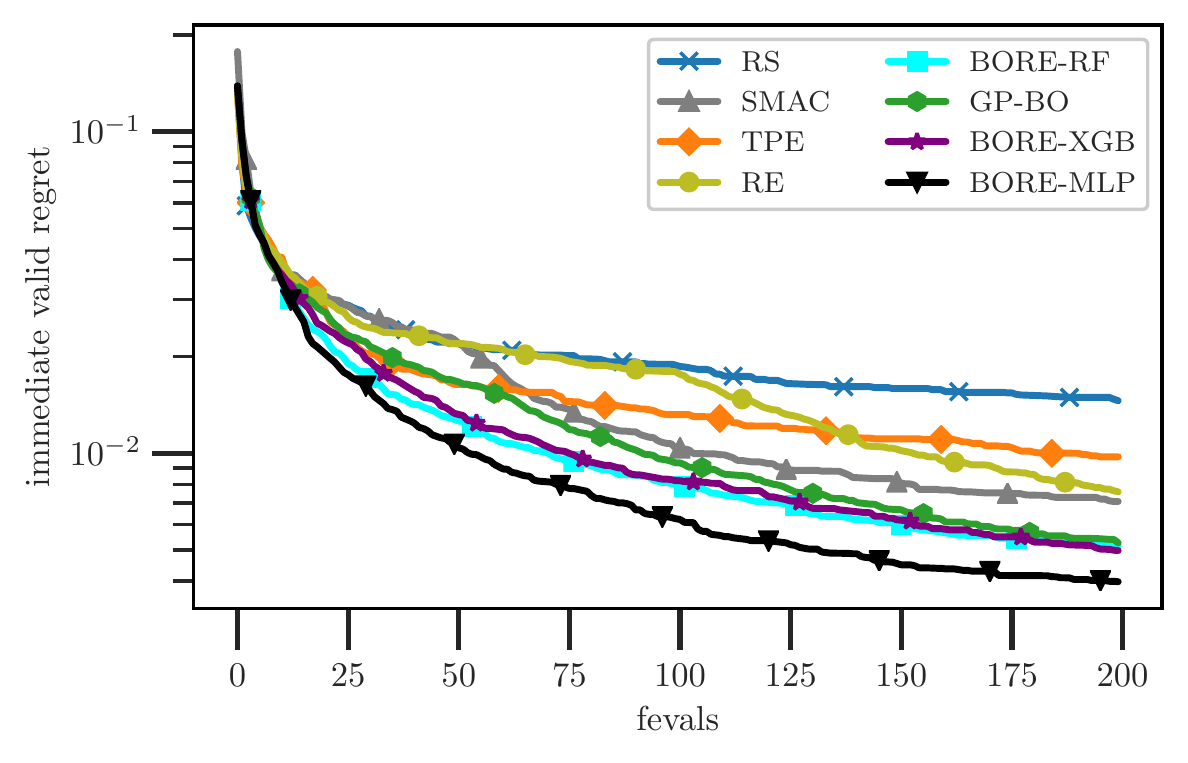}
    \caption{ImageNet-16}
  \end{subfigure}
   ~
  \caption{Immediate regret over function evaluations on the NASBench201 \acrlong{NAS} problems ($D=6$).}
  \label{fig:nasbench201}
\end{figure*}
\parhead{Neural architecture search (NASBench201).} 
Next, we consider a \gls{NAS} problem, namely, that of designing a neural cell.
A cell is represented by a \gls{DAG} with 4 nodes, and the task is to assign an 
\emph{operation} to each of the 6 
possible arcs 
from
a set of five operations. 
\LT{If space becomes a problem: make this more concise or less specific; defer details to appendix}
We utilize NASBench201~\cite{dong2020bench}, which
tabulates 
precomputed results from 
all possible $5^6=15,625$ combinations for each of the three datasets: 
CIFAR-10, CIFAR-100~\cite{krizhevsky2009learning} and ImageNet-16~\cite{chrabaszcz2017downsampled}.
Additional details are included in \cref{sub:nasbench},
and the results are shown in \cref{fig:nasbench201}.
We find across all datasets that the \gls{BORE} variants consistently achieve 
the lowest final regret among all baselines. Not only that, the \gls{BORE} 
variants, in particular \acrshort{BOREMLP}, maintains the lowest regret at 
anytime (i.e. 
at any optimization iteration), followed by \acrshort{BORERF}, then \acrshort{BOREXGB}.
In this problem, the inputs are purely categorical, whereas in the previous problem
they are a mix of categorical and ordinal.
For the \acrshort{BOREMLP} variant, categorical inputs are one-hot encoded, 
while ordinal inputs are handled by simply rounding to its nearest integer 
index. 
The latter is known to have shortcomings~\cite{garrido2020dealing}, and might 
explain why \acrshort{BOREMLP} is the most effective variant in this problem 
but the least effective in the previous one.


\begin{figure}[ht]
  \centering
  \includegraphics[width=\linewidth]{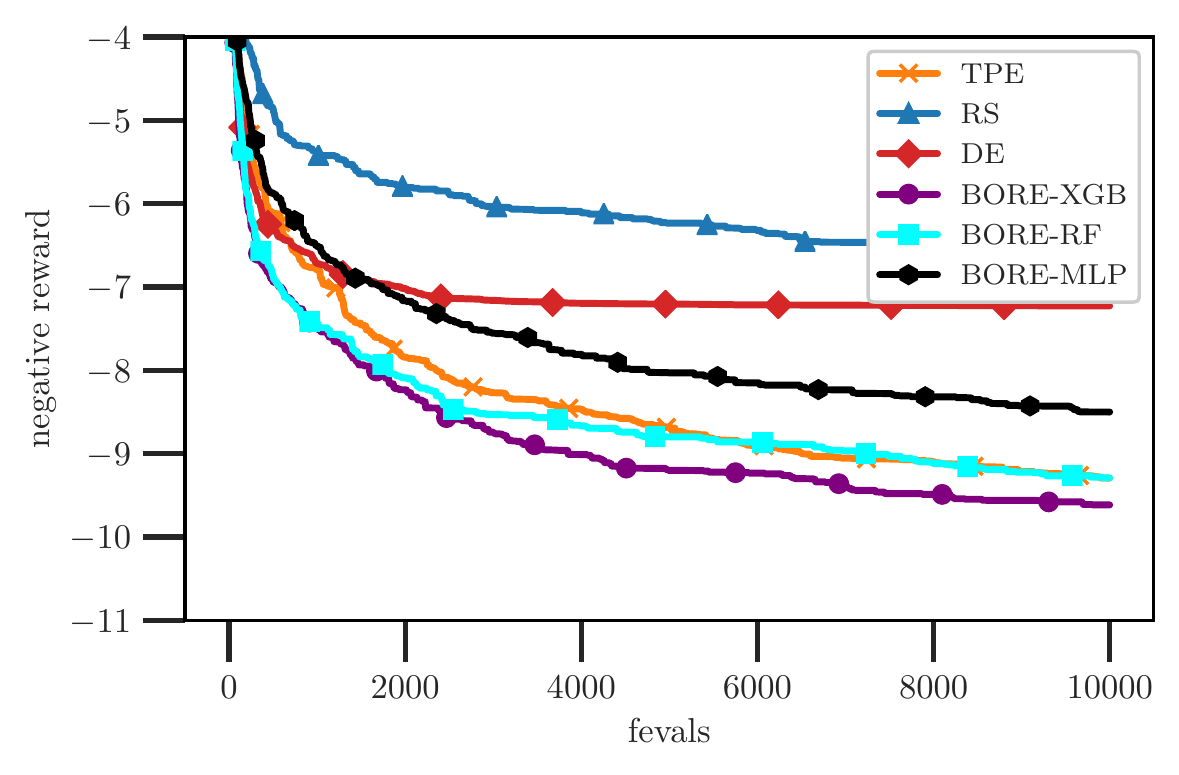}
  \vskip -0.1in
  \caption{Negative reward over function evaluations on the Robot Pushing task ($D=14$).}
  \label{fig:robot_pushing}
  \vskip -0.1in
\end{figure}
\parhead{Robot arm pushing.}
We consider the 14D control problem first studied by \citet{wang2017max}.
The problem is concerned with tuning the controllers of robot hands to push objects to some desired locations. 
Specifically, there are two robots, each tasked with manipulating an object. 
For each robot, the control parameters include the \emph{location} and 
\emph{orientation} of its hands, the \emph{moving direction}, \emph{pushing speed}, and \emph{duration}.
Due to the large number of function evaluations~($\sim$10,000) required to 
achieve a reasonable performance, we omit \textsc{gp-bo} from our comparisons 
on this benchmark. 
Further, we reduce the number of replicated runs of each method to 50. 
Additional details are included in \cref{sub:robot_pushing_control},
and the results are shown in \cref{fig:robot_pushing}.
We see that \acrshort{BOREXGB} attains the highest reward, followed by 
\acrshort{BORERF} and \gls{TPE} (which attain roughly the same performance), and then \acrshort{BOREMLP}. 

\begin{figure*}[ht]
  \centering
  \begin{subfigure}[t]{0.32\textwidth}
    \centering
    \includegraphics[width=\linewidth]{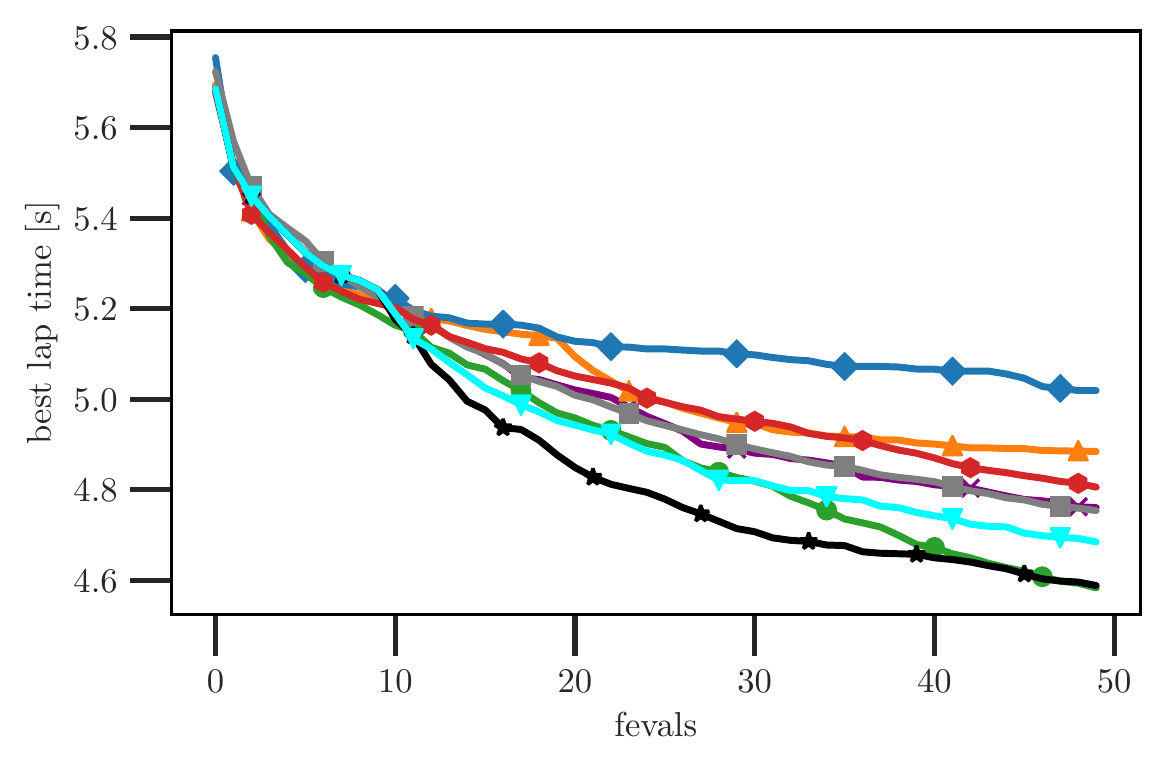}
    \caption{\textsc{uc berkeley} ($D=12$)}
  \end{subfigure}
  ~ 
  \begin{subfigure}[t]{0.32\textwidth}
    \centering
    \includegraphics[width=\linewidth]{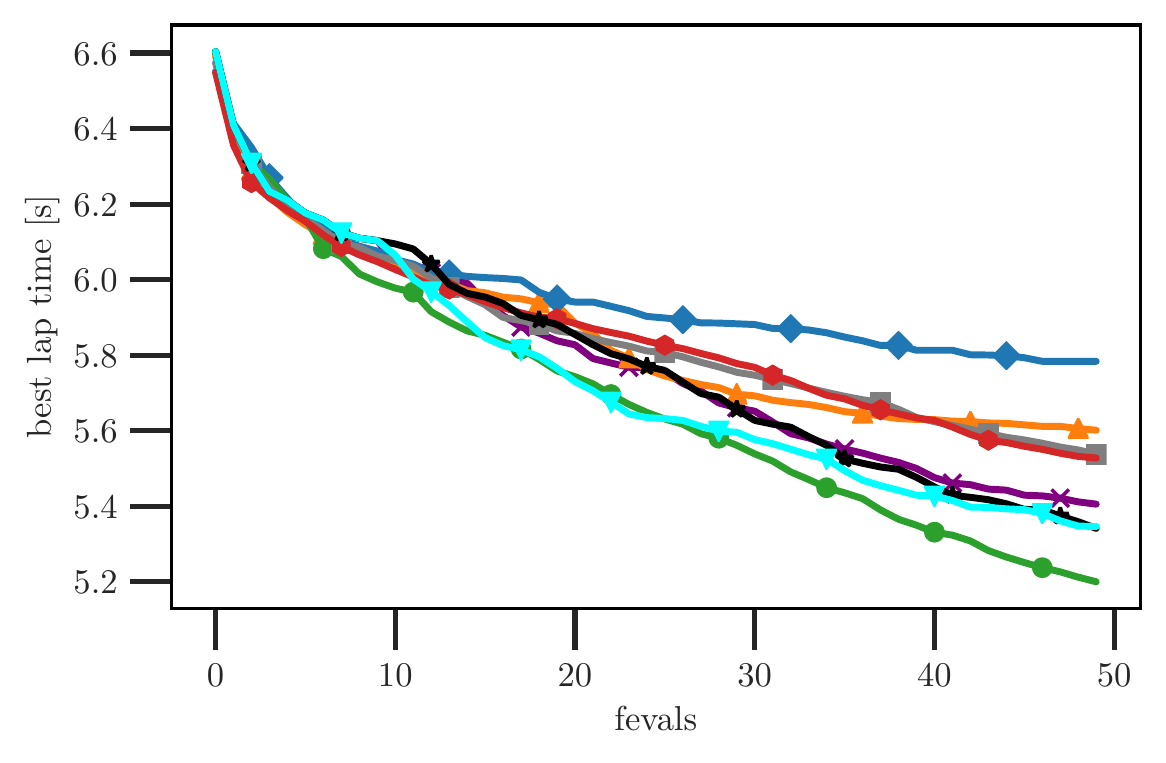}
    \caption{\textsc{eth z\"{u}rich} A ($D=20$)}
  \end{subfigure}
  ~
  \begin{subfigure}[t]{0.32\textwidth}
    \centering
    \includegraphics[width=\linewidth]{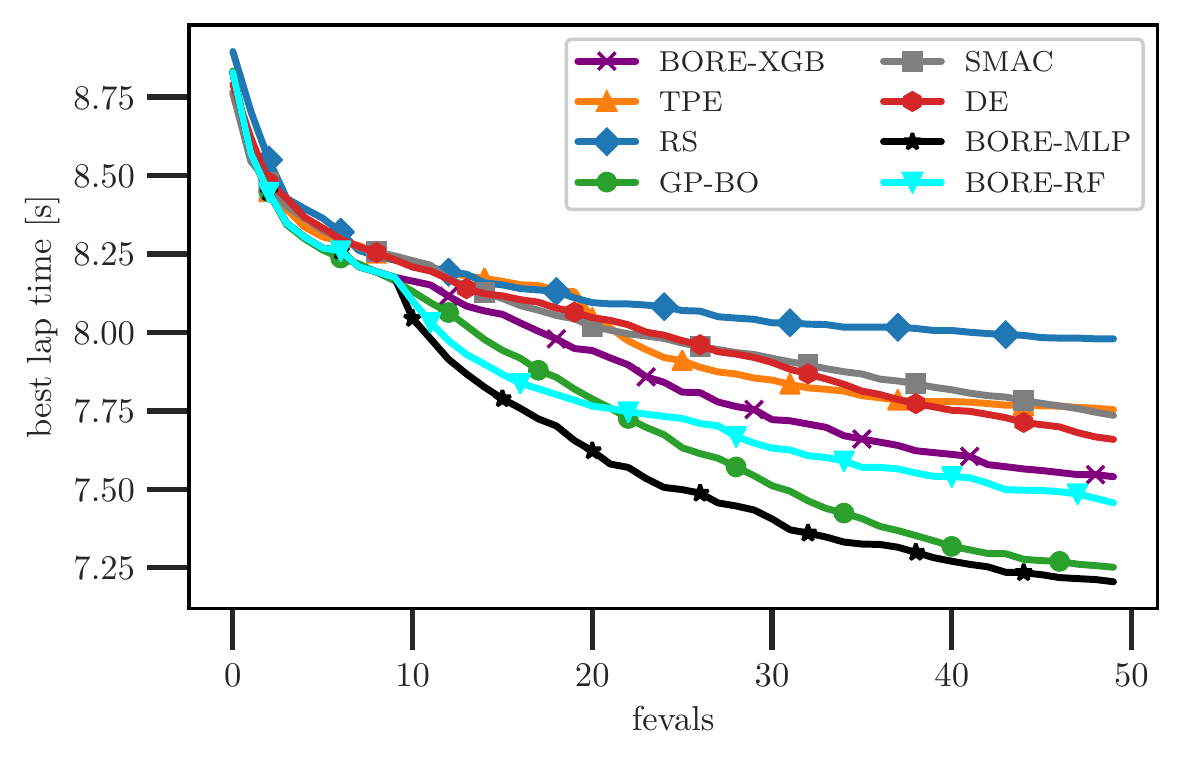}
    \caption{\textsc{eth z\"{u}rich} B ($D=21$)}
  \end{subfigure}
  \caption{Best lap times (in seconds) over function evaluations in the racing line optimization problem on various racetracks.}
  \label{fig:racing}
\end{figure*}
\parhead{Racing line optimization.}
We consider the problem of computing the optimal racing line for a given 
track and vehicle with known dynamics.
We adopt the set-up of~\citet{jain2020computing},
who consider the dynamics of miniature scale 
cars traversing 
around tracks at \textsc{uc berkeley} and \textsc{eth z\"{u}rich}.
The racing line is a trajectory determined by $D$ waypoints placed along the 
length of the track, where the $i$th waypoint deviates from the centerline of 
the track by $x_i \in \left [ - \frac{W}{2}, \frac{W}{2} \right ]$ for some 
track width $W$.
The task is to minimize  the lap time $f(\mbx)$, the
minimum 
time required 
to traverse the trajectory parameterized by 
$\mbx = [ x_1 \cdots x_D ]^{\top}$. 
Additional details are included in \cref{sub:racing_line_optimization},
and the results are shown in \cref{fig:racing}.
First, we see that the \gls{BORE} variants consistently outperform all 
baselines except for \textsc{gp-bo}.
This is to be expected, since the function is continuous, smooth and has 
$\sim$20 dimensions or less.
Nonetheless, we find that the \acrshort{BOREMLP} variant performs as well as, or 
marginally better than, \textsc{gp-bo} on two tracks.
In particular, on the \textsc{uc berkeley} track, we see that \acrshort{BOREMLP} 
achieves the best lap times for the first $\sim$40 evaluations, and is caught 
up to by \textsc{gp-bo} in the final 10. 
On \textsc{eth z\"{u}rich} track \textsc{b}, \acrshort{BOREMLP} consistently maintains a narrow lead.

\begin{figure}[ht]
  \centering
  \includegraphics[width=\linewidth]{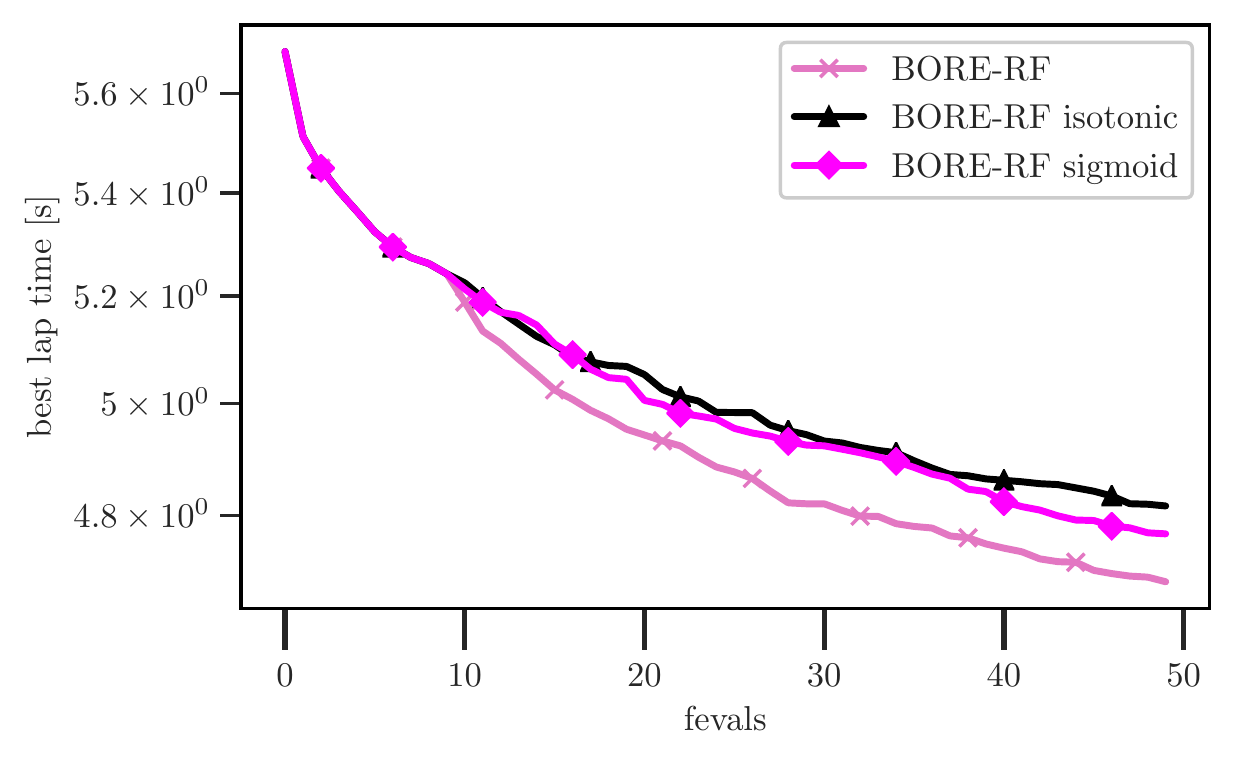}
  \vskip -0.1in
  \caption{Effects of calibrating \glspl{RF} in the \acrshort{BORERF} variant.
           Results of racing line optimization on the \textsc{uc berkeley} track.}
  \label{fig:calibration-racing}
  \vskip -0.1in
\end{figure}
{\it Effects of calibration.}
As discussed in~\cref{sub:choice_of_probabilistic_classifier},
calibrating \glspl{RF} may have a profound effect on the \acrshort{BORERF} variant.
We consider two popular approaches~\cite{niculescu2005predicting}, namely,
Platt scaling~\cite{platt1999probabilistic} and
isotonic regression~\cite{zadrozny2001obtaining,zadrozny2002transforming}.
The results shown in \cref{fig:calibration-racing} suggest that 
applying these calibration techniques may have deleterious effects. 
However, this 
can also be adequately
explained by overfitting due to 
insufficient calibration samples (in the case of isotonic regression, 
$\sim$1,000 samples are necessary). 
Therefore, we may yet observe the benefits of calibration in 
problem
settings
that 
yield large amounts of data. 

We provide further ablation studies in~\cref{sec:ablation_studies}.

\section{Discussion and Outlook}
\label{sec:discussion_outlook}
We examine the 
limitations of our method, 
discuss how these may be addressed,
and outline 
additional future directions.

We restricted our attention to a simple treatment of hyperparameters $\mbtheta$
based on point estimates. 
For example, in the \acrshort{BOREMLP} variant this consists of the weights and biases. 
For improved exploration, it may be beneficial to consider placing a prior on $\mbtheta$ 
and marginalizing out its uncertainty~\cite{snoek2012practical}.
Refer to \cref{sub:hyperparameters} for an expanded discussion. 
Further, compared against \textsc{gp-bo}, a potential downside of \gls{BORE} is 
that there may be vastly more \emph{meta-hyperparameters} settings from which to choose.
Whereas in \textsc{gp-bo} these might consist of, e.g. the choice of kernel 
and 
its isotropy, 
there are potentially many more possibilities in \gls{BORE}.
In the case of \acrshort{BOREMLP}, this may consist of, e.g. layer depth, widths, 
activations, etc---the tuning of which is often the reason one appeals to 
\gls{BO} in the first place.
While we obtained remarkable results with the proposed variants without
needing to
deviate from the sensible defaults,  
generally speaking,
for further improvements in calibration and sample diversity, it may be 
beneficial to consider hyper-deep ensembles~\cite{wenzel2020hyperparameter}.
Refer to \cref{sub:meta_hyperparameters} for further discussion. 

Another avenue to explore is the potential benefits of other direct \gls{DRE} methods,
in particular \gls{RULSIF}~\cite{yamada2011relative}, which is the only 
method of those aforementioned in \cref{sec:related_work} that directly 
estimates the \emph{relative} density-ratio. 
Furthermore, since \gls{RULSIF} is parameterized by a 
sum of Gaussian kernels,
it enables the use of well-established mode-finding approaches, 
such as the \emph{mean-shift} 
algorithm~\cite{comaniciu2002mean},
for candidate suggestion.
Along the same avenue, but in 
a different direction, one may also consider employing \gls{DRE} losses for 
classifier learning~\cite{menon2016linking}.

Lastly, a fertile ground for exploration is extending
\gls{BORE} with classifier designs suitable for \gls{BO} more sophisticated paradigms, 
such as in the 
multi-task~\cite{swersky2013multi}, 
multi-fidelity~\cite{kandasamy2017multi}, and 
multi-objective settings~\cite{hernandez2016predictive}, 
in addition to
architectures effective for \gls{BO} of
sequences~\cite{moss2020boss} and by extension, 
molecular structures~\cite{gomez2018automatic} and beyond.


\section{Conclusion}
\label{sec:conclusion}

We have presented a novel methodology for \gls{BO} 
based on the 
observation
that the problem of computing 
\gls{EI} 
can be reduced to that of probabilistic classification.
This 
observation
is made through the well-known link between \gls{CPE} and \gls{DRE}, 
and the lesser-known insight that \gls{EI} can be expressed as a relative 
density-ratio between two unknown distributions.

We discussed important ways in which \gls{TPE}, an early attempt to exploit the 
latter link,
falls short.
Further, we demonstrated that our \gls{CPE}-based approach to \gls{BORE},
in particular, our variants based on the \gls{MLP}, \gls{RF} and \gls{XGBOOST} classifiers,
consistently outperform \gls{TPE}, and 
compete well against
the state-of-the-art 
derivative-free global optimization methods.

Overall, the simplicity and effectiveness of \gls{BORE} make it a 
promising approach for blackbox optimization, and its high degree of 
extensibility provides numerous exciting avenues for future work.




\clearpage

\bibliography{main}
\bibliographystyle{icml2021}

\onecolumn

\appendix

\glsresetall
\glsunset{BORE}

\section{Expected improvement}
\label{sec:derivation}
For completeness, we reproduce the derivations of \citet{bergstra2011algorithms}.
Recall from \cref{eq:expected-improvement-generic} that the \gls{EI} 
function is defined as the expectation of the improvement utility function 
$U(\mbx, y, \tau)$ over the posterior predictive distribution $p(y \g \mbx, \cD_N)$.
Expanding this out, we have
\begin{align*}
  \alpha(\mbx; \cD_N, \tau) \defeq \bbE_{p(y \g \mbx, \cD_N)}[U(\mbx, y, \tau)] 
  & = \int_{-\infty}^{\infty} U(\mbx, y, \tau) p(y \g \mbx, \cD_N) \, \mathrm{d}y \\
  & = \int_{-\infty}^{\tau} (\tau - y) p(y \g \mbx, \cD_N) \, \mathrm{d}y \\
  & = \frac{1}{p(\mbx \g \cD_N)} \int_{-\infty}^{\tau} (\tau - y) p(\mbx \g y, \cD_N) p(y \g \cD_N) \, \mathrm{d}y, \label{eq:bayes1}
\end{align*}
where we have invoked Bayes' rule in the final step above.
Next, the denominator evaluates to
\begin{align*}
  p(\mbx \g \cD_N) 
  & = \int_{-\infty}^{\infty} p(\mbx \g y, \cD_N) p(y \g \cD_N) \, \mathrm{d}y \\
  & = \ell(\mbx) \int_{-\infty}^{\tau} p(y \g \cD_N) \, \mathrm{d}y + g(\mbx) \int_{\tau}^{\infty} p(y \g \cD_N) \, \mathrm{d}y \\
  & = \gamma \ell(\mbx) + (1 - \gamma) g(\mbx),
\end{align*}
since, by definition, $\gamma = \Phi(\tau) \defeq p(y \leq \tau \g \cD_N)$.
Finally, we evaluate the numerator,
\begin{align*}
  \int_{-\infty}^{\tau} (\tau - y) p(\mbx \g y, \cD_N) p(y \g \cD_N) \, \mathrm{d}y
  & = \ell(\mbx) \int_{-\infty}^{\tau} (\tau - y) p(y \g \cD_N) \, \mathrm{d}y \\
  & = \ell(\mbx) \tau \int_{-\infty}^{\tau} p(y \g \cD_N) \, \mathrm{d}y - \ell(\mbx) \int_{-\infty}^{\tau} y p(y \g \cD_N) \, \mathrm{d}y \\
  & = \gamma \tau \ell(\mbx) - \ell(\mbx) \int_{-\infty}^{\tau} y p(y \g \cD_N) \, \mathrm{d}y \\
  & = K \cdot \ell(\mbx),
\end{align*}
where
\begin{equation*}  
  K = \gamma \tau - \int_{-\infty}^{\tau} y p(y \g \cD_N) \, \mathrm{d}y.
\end{equation*}

Hence, this shows that the \gls{EI} function is equivalent to the 
$\gamma$-relative density ratio \citep{yamada2011relative} up to a constant 
factor $K$,
\begin{align*}
  \alpha(\mbx; \cD_N, \tau)
  & \propto
  \frac{\ell(\mbx)}{\gamma \ell(\mbx) + (1 - \gamma) g(\mbx)} \\
  & = \left ( \gamma + \frac{g(\mbx)}{\ell(\mbx)} (1 - \gamma) \right )^{-1}.
\end{align*}

\section{Relative density-ratio: unabridged notation}
\label{sec:relative_density_ratio}

In \cref{sec:background}, for notational simplicity, we had excluded the 
dependencies of $\ell, g$ and $r_{\gamma}$ on $\tau$.
Let us now define these densities more explicitly as
\begin{equation*}
  \ell(\mbx; \tau) \defeq p \left (\mbx \g y \leq \tau, \cD_N \right ),
  \qquad
  \text{and}
  \qquad  
  g(\mbx; \tau) \defeq p \left (\mbx \g y > \tau, \cD_N \right ),
\end{equation*}
and accordingly, the $\gamma$-relative density-ratio from \cref{eq:density-ratio-relative} as
\begin{equation*}
  r(\mbx; \gamma,\tau)
  = \frac{\ell(\mbx; \tau)}
         {\gamma \ell(\mbx; \tau) + (1-\gamma) g(\mbx; \tau)}.
\end{equation*}
Recall from \cref{eq:ei-as-relative-density-ratio} that
\begin{equation} \label{eq:ei-as-relative-density-ratio-tau}
  \alpha \left (\mbx; \cD_N , \Phi^{-1}(\blue{\gamma}) \right ) 
  \propto 
  r \left (\mbx; \blue{\gamma}, \Phi^{-1}(\blue{\gamma}) \right ).  
\end{equation}
In \cref{itm:ordinary}, \citet{bergstra2011algorithms} resort to optimizing~$r(\mbx; 0, \Phi^{-1}(\gamma))$,
which is justified by the fact that
\begin{equation*}
  r(\mbx; \blue{\gamma}, \Phi^{-1}(\blue{\gamma})) 
  =
  h_{\blue{\gamma}} \left [ r \left (\mbx; \orange{0}, \Phi^{-1}(\blue{\gamma}) \right ) \right ],
\end{equation*}
for strictly nondecreasing $h_{\gamma}$.
Note we have used a \blue{blue} and \orange{orange} color coding to emphasize 
the differences in the setting of~$\gamma$ (best viewed on a computer screen).
Recall that $\Phi^{-1}(0) = \min_n y_n$ corresponds to the conventional 
setting of threshold $\tau$.
However, make no mistake,
for any $\gamma > 0$,
\begin{equation*}
  \alpha \left (\mbx; \cD_N , \Phi^{-1}(\orange{0}) \right ) 
  \not \propto 
  r \left (\mbx; \orange{0}, \Phi^{-1}(\blue{\gamma}) \right ).
\end{equation*}
Therefore, given the numerical instabilities associated with this approach
as discussed in \cref{sub:potential_pitfalls}, 
there is no advantage to be gained from taking this direction.
Moreover, \cref{eq:ei-as-relative-density-ratio-tau} only holds for $\gamma > 0$.
To see this, suppose $\gamma = 0$, which gives
\begin{equation*}
  \alpha \left (\mbx; \cD_N , \Phi^{-1}(\blue{0}) \right ) 
  \propto 
  r \left (\mbx; \blue{0}, \Phi^{-1}(\blue{0}) \right ).
\end{equation*}
However, since by definition $\ell \left (\mbx; \Phi^{-1}(0) \right )$ has no mass,
the \textsc{rhs} is undefined.

\section{Class-posterior probability}
\label{sec:class_posterior_probability}

We provide an unabridged derivation of the identity of \cref{eq:relative-density-ratio-class-posterior}.
First, the $\gamma$-relative density ratio is given by
\begin{align*}
  r_{\gamma}(\mbx) 
  & \defeq \frac{\ell(\mbx)}{\gamma \ell(\mbx) + (1 - \gamma) g(\mbx)} \\
  & = \frac{p(\mbx \g z = 1)}{\gamma \cdot p(\mbx \g z = 1) + (1 - \gamma) \cdot p(\mbx \g z = 0)} \\
  & = \left ( \frac{p(z = 1 \g \mbx) \cancel{p(\mbx)}}{p(z = 1)} \right )
      \left ( \gamma \cdot \frac{p(z = 1 \g \mbx) \cancel{p(\mbx)}}{p(z = 1)} + 
        (1 - \gamma) \cdot \frac{p(z = 0 \g \mbx) \cancel{p(\mbx)}}{p(z = 0)} \right )^{-1}.
\end{align*}
By construction, we have $p(z = 1) \defeq p(y \leq \tau) = \gamma$ and 
$\pi(\mbx) \defeq p(z = 1 \g \mbx)$.
Therefore,
\begin{align*}
  r_{\gamma}(\mbx) 
  & = \gamma^{-1} \pi(\mbx)
      \left ( \cancel{\gamma} \cdot \frac{\pi(\mbx) }{\cancel{\gamma}} + 
              \cancel{(1 - \gamma)} \cdot \frac{1 - \pi(\mbx)}{ \cancel{1 - \gamma}} \right )^{-1} \\
  & = \gamma^{-1} \pi(\mbx).
\end{align*}
Alternatively, we can also arrive at the same result by writing the ordinary 
density ratio $r_0(\mbx)$ in terms of $\pi(\mbx)$ and $\gamma$, which is 
well-known to be
\begin{equation*}
  r_0(\mbx)
  = \left ( \frac{\gamma}{1 - \gamma} \right )^{-1} \frac{\pi(\mbx)}{1 - \pi(\mbx)}.
\end{equation*}
Plugging this into function $h_{\gamma}$, we get
\begin{align*}
  r_{\gamma}(\mbx) = h_{\gamma}(r_0(\mbx))
  & = h_{\gamma}\left ( \left ( \frac{\gamma}{1 - \gamma} \right )^{-1} \frac{\pi(\mbx)}{1 - \pi(\mbx)} \right ) \\
  & = \left ( \gamma + (1 - \gamma) \left ( \frac{\gamma}{1 - \gamma} \right ) \left ( \frac{\pi(\mbx)}{1 - \pi(\mbx)} \right )^{-1}  \right )^{-1} \\
  & = \gamma^{-1} \left ( 1 + \left ( \frac{\pi(\mbx)}{1 - \pi(\mbx)} \right )^{-1} \right )^{-1} \\
  & = \gamma^{-1} \pi(\mbx).
\end{align*}

\section{Log loss}
\label{sec:log_loss}

The log loss, also known as the \gls{BCE} loss, is given by
\begin{equation} \label{eq:log-loss}
  \cL^*(\mbtheta) 
  \defeq 
  - \beta \cdot \bbE_{\ell(\mbx)} [\log{\pi_\mbtheta(\mbx)}] 
  - (1-\beta) \cdot \bbE_{g(\mbx)} [\log{(1-\pi_\mbtheta(\mbx))}],
\end{equation}
where $\beta$ denotes the class balance rate. 
In particular, let $N_{\ell}$ and $N_{g}$ be the sizes of the support 
of $\ell(\mbx)$ and $g(\mbx)$, respectively.
Then, we have
\begin{equation*}
  \beta = \frac{N_{\ell}}{N},
  \qquad
  \text{and}
  \qquad
  1 - \beta = \frac{N_{g}}{N},
\end{equation*}
where $N = N_{\ell} + N_{g}$.
In practice, we approximate the log loss $\cL^*(\mbtheta)$ by the empirical 
risk of \cref{eq:log-loss-empirical}, given by
\begin{equation*} 
  \cL(\mbtheta)
  \defeq
  - \frac{1}{N} \left ( \sum_{n=1}^N 
    z_n \log{\pi_\mbtheta(\mbx_n)} + 
    (1 - z_n) \log{(1-\pi_\mbtheta(\mbx_n))} \right ).
\end{equation*}
In this section, we show that the approximation of \cref{eq:approximation}, 
that is,
\begin{equation*}
  \pi_\mbtheta(\mbx) \simeq \gamma \cdot r_{\gamma}(\mbx),
\end{equation*}
attains
equality 
at $\mbtheta_{\star} = \argmin_{\mbtheta} \cL^*(\mbtheta)$. 

\subsection{Optimum}
\label{sub:optimum}

Taking the functional derivative of $\cL^*$ in \cref{eq:log-loss}, 
we get
\begin{align*}
  \frac{\partial \cL^*}{\partial \pi_\mbtheta}
  & = 
  - 
  \bbE_{\ell(\mbx)} \left [ 
    \frac{\beta}{\pi_\mbtheta(\mbx)}
  \right ] 
  + 
  \bbE_{g(\mbx)} \left [ 
    \frac{1 - \beta}{1- \pi_\mbtheta(\mbx)}
  \right ] \\
  & = 
  \int
  \left ( 
  - \beta \frac{\ell(\mbx)}{\pi_\mbtheta(\mbx)}
  + (1 - \beta) \frac{g(\mbx)}{1- \pi_\mbtheta(\mbx)}
  \right )
  \mathrm{d}\mbx
\end{align*}
This integral evaluates to zero iff the integrand itself evaluates to zero.
Hence, we solve the following for $\pi_{\mbtheta_{\star}}(\mbx)$,
\begin{equation*}
  \beta  \frac{\ell(\mbx)}{\pi_{\mbtheta_{\star}}(\mbx)}
  = 
  (1 - \beta) \frac{g(\mbx) }{1 - \pi_{\mbtheta_{\star}}(\mbx)}.
\end{equation*}
We re-arrange this expression to give
\begin{equation*}
  \frac{1- \pi_{\mbtheta_{\star}}(\mbx)}{\pi_{\mbtheta_{\star}}(\mbx)}
  = 
  \left ( 
    \frac{1-\beta}{\beta}
  \right ) \frac{g(\mbx)}{\ell(\mbx)} 
  \qquad
  \Leftrightarrow
  \qquad
  \frac{1}{\pi_{\mbtheta_{\star}}(\mbx)} - 1
  = 
  \frac{\beta \ell(\mbx) + (1-\beta) g(\mbx)}{\beta \ell(\mbx)} - 1.
\end{equation*}
Finally, we add one to both sides and invert the result to give
\begin{align*}
  \pi_{\mbtheta_{\star}}(\mbx)
  & = 
  \frac{\beta \ell(\mbx)}{\beta \ell(\mbx) + (1-\beta) g(\mbx)} \\
  & = \beta \cdot r_{\beta}(\mbx).
\end{align*}
Since, by definition $\beta = \gamma$, this leads to
$\pi_{\mbtheta_{\star}}(\mbx) = \gamma \cdot r_{\gamma}(\mbx)$ as required.

\subsection{Empirical risk minimization}
\label{sub:empirical_risk_minimization}

For completeness, we show that the log loss $\cL^*(\mbtheta)$ of \cref{eq:log-loss} 
can be approximated by $\cL(\mbtheta)$ of \cref{eq:log-loss-empirical}.
First, let $\rho$ be the permutation of the set $\{1, \dotsc, N\}$, i.e. the bijection 
from $\{1, \dotsc, N\}$ to itself, such that $y_{\rho(n)} \leq \tau$ if 
$0 < \rho(n) \leq N_{\ell}$, and $y_{\rho(n)} > \tau$ if $N_{\ell} < \rho(n) \leq N_{g}$.
That is to say,
\begin{equation*}
  \mbx_{\rho(n)} \sim
  \begin{cases} 
    \ell(\mbx) & \text{if } 0 < \rho(n) \leq N_{\ell}, \\
    g(\mbx)    & \text{if } N_{\ell} < \rho(n) \leq N_{g}.
  \end{cases}
  \quad
  \text{and}
  \quad
  z_{\rho(n)} \defeq
  \begin{cases} 
    1 & \text{if } 0 < \rho(n) \leq N_{\ell}, \\
    0 & \text{if } N_{\ell} < \rho(n) \leq N_{g}.
  \end{cases}
\end{equation*}
Then, we have
\begin{align*}
  \cL^*(\mbtheta) 
  & \defeq 
  - \frac{1}{N} 
  \left ( N_{\ell} \cdot \bbE_{\ell(\mbx)} [\log{\pi_\mbtheta(\mbx)}] 
  + N_{g} \cdot \bbE_{g(\mbx)} [\log{(1-\pi_\mbtheta(\mbx))}] \right ) \\
  & \simeq 
  - \frac{1}{N} 
  \left ( \cancel{N_{\ell}} \cdot \frac{1}{\cancel{N_{\ell}}} \sum_{n=1}^{N_{\ell}} \log{\pi_\mbtheta(\mbx_{\rho(n)})}
  + \cancel{N_{g}} \cdot \frac{1}{\cancel{N_{g}}} \sum_{n=N_{\ell}+1}^{N_{g}} \log{(1-\pi_\mbtheta(\mbx_{\rho(n)}))} \right ) \\
  & =
  - \frac{1}{N}
  \left ( \sum_{n=1}^{N} z_{\rho(n)} \log{\pi_\mbtheta(\mbx_{\rho(n)})}
                       + (1 - z_{\rho(n)}) \log{(1-\pi_\mbtheta(\mbx_{\rho(n)}))} \right ) \\
  & =
  - \frac{1}{N}
  \left ( \sum_{n=1}^{N} z_n \log{\pi_\mbtheta(\mbx_n)}
                       + (1 - z_n) \log{(1-\pi_\mbtheta(\mbx_n))} \right )
  \defeq 
  \cL(\mbtheta),
\end{align*}
as required.

\section{Step-through visualization} 
\label{sec:step_through_visualization}

\begin{figure}[H]
  \centering
  \begin{subfigure}[t]{0.49\textwidth}
    \centering
    \includegraphics[width=\columnwidth]{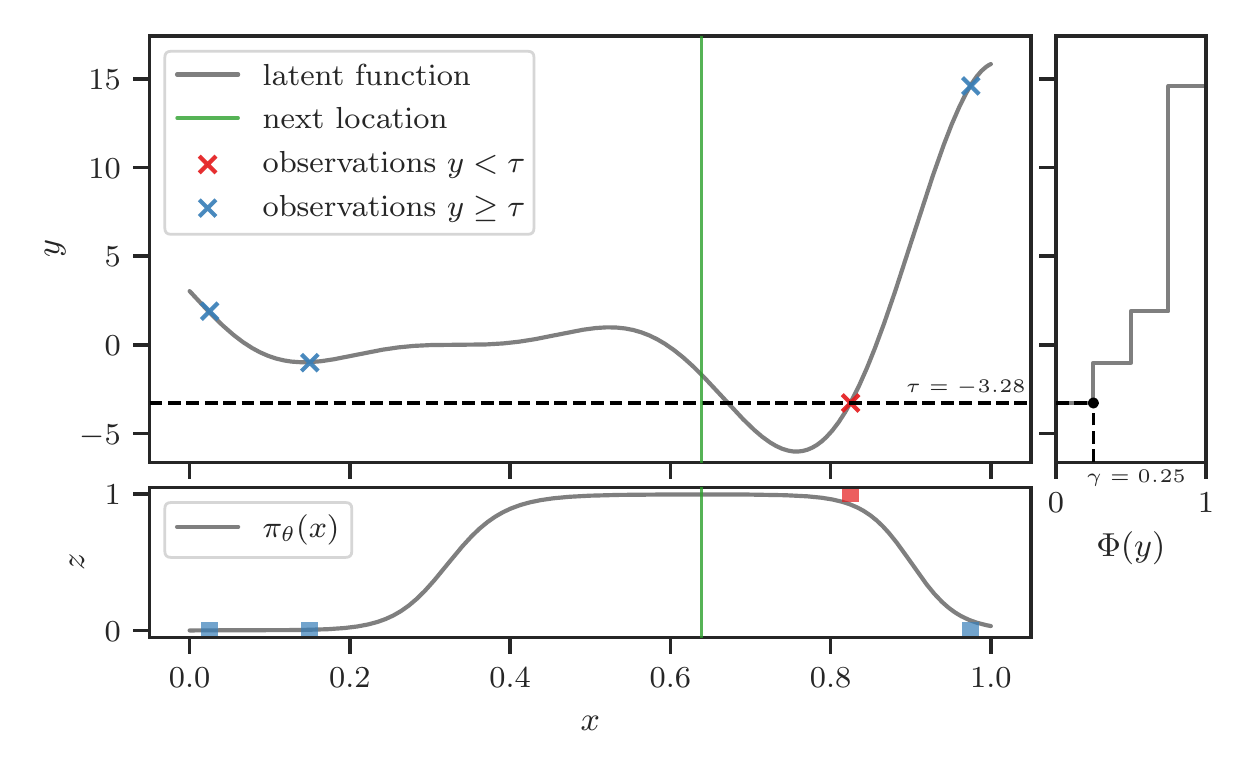}
    \caption{Iteration 1}
  \end{subfigure}
  ~
  \begin{subfigure}[t]{0.49\textwidth}
    \centering
    \includegraphics[width=\columnwidth]{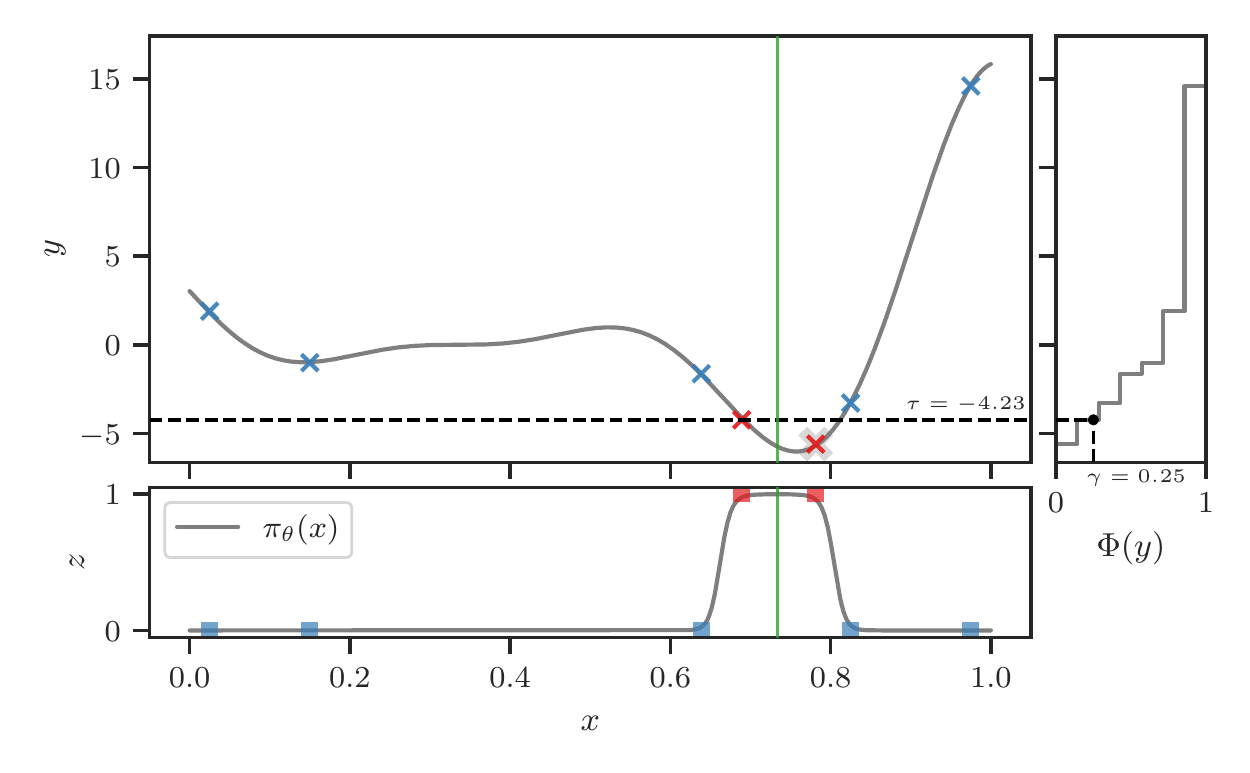}
    \caption{Iteration 4}
  \end{subfigure}
  \begin{subfigure}[t]{0.49\textwidth}
    \centering
    \includegraphics[width=\columnwidth]{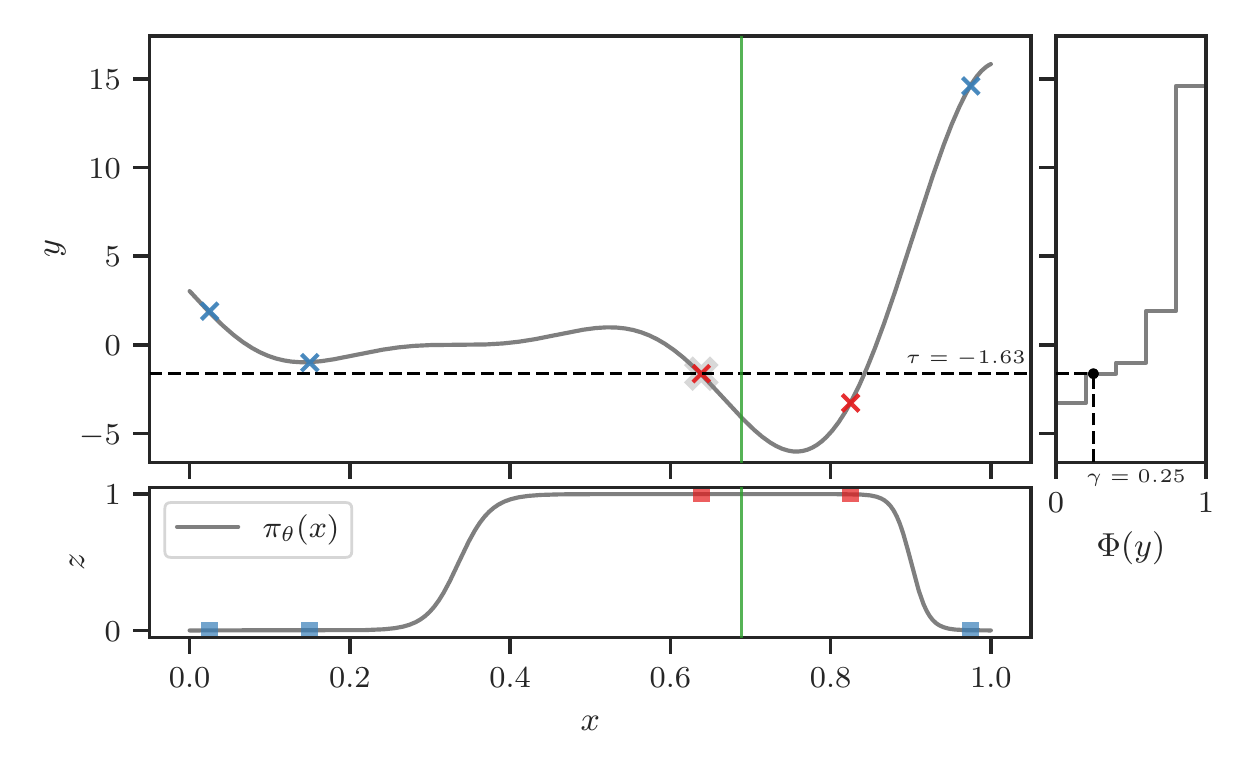}
    \caption{Iteration 2}
  \end{subfigure}
  ~
  \begin{subfigure}[t]{0.49\textwidth}
    \centering
    \includegraphics[width=\columnwidth]{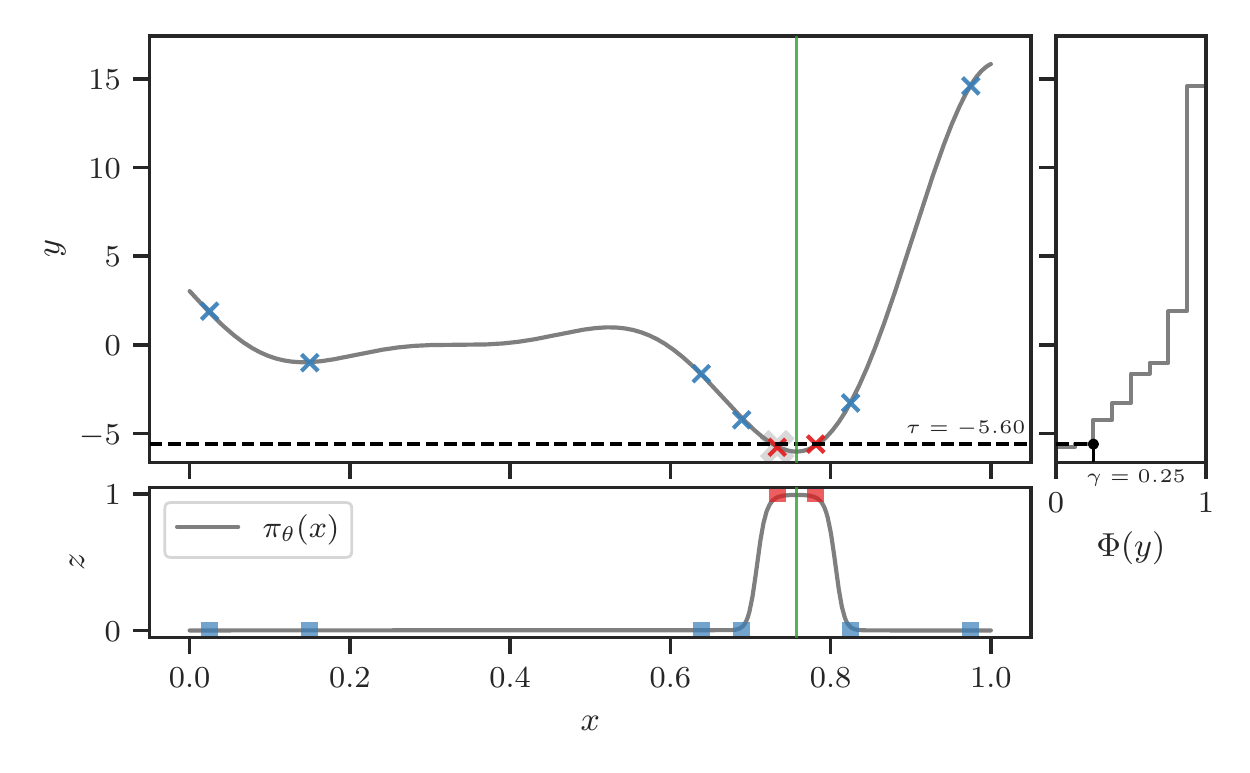}
    \caption{Iteration 5}
  \end{subfigure}
  \begin{subfigure}[t]{0.49\textwidth}
    \centering
    \includegraphics[width=\columnwidth]{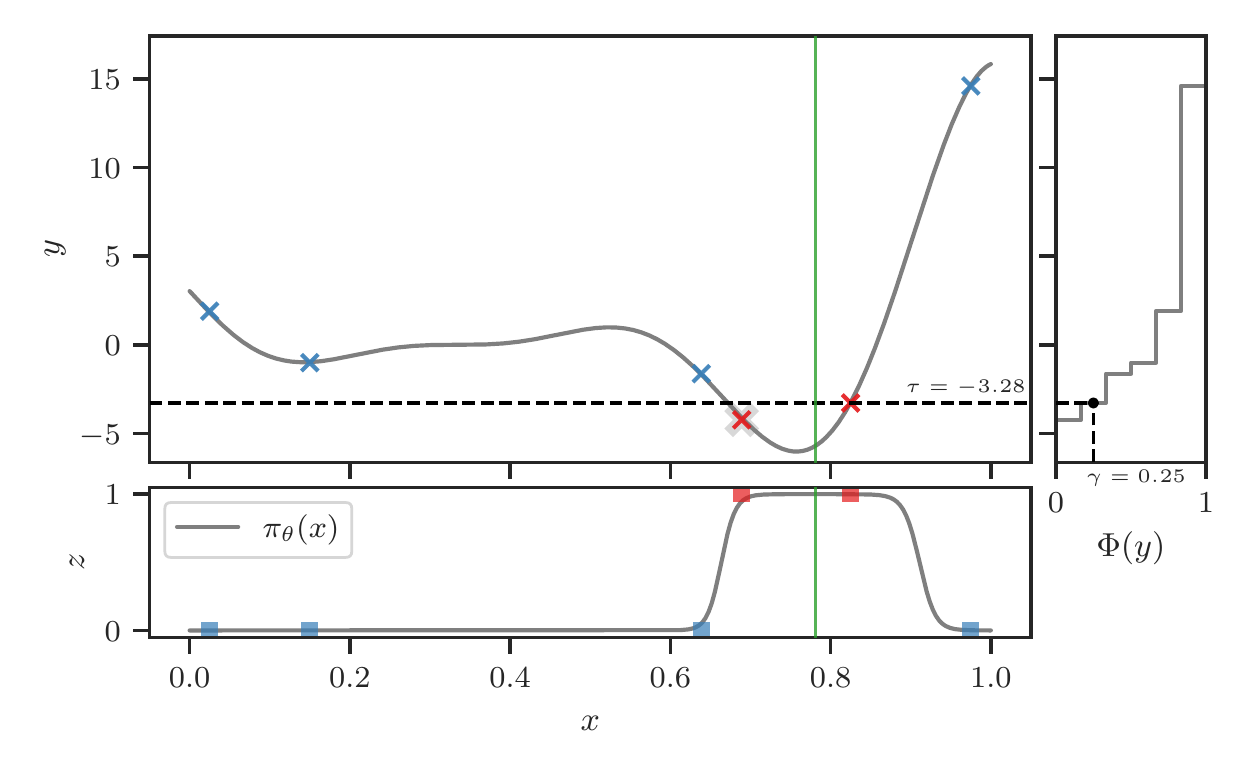}
    \caption{Iteration 3}
  \end{subfigure}
  ~
  \begin{subfigure}[t]{0.49\textwidth}
    \centering
    \includegraphics[width=\columnwidth]{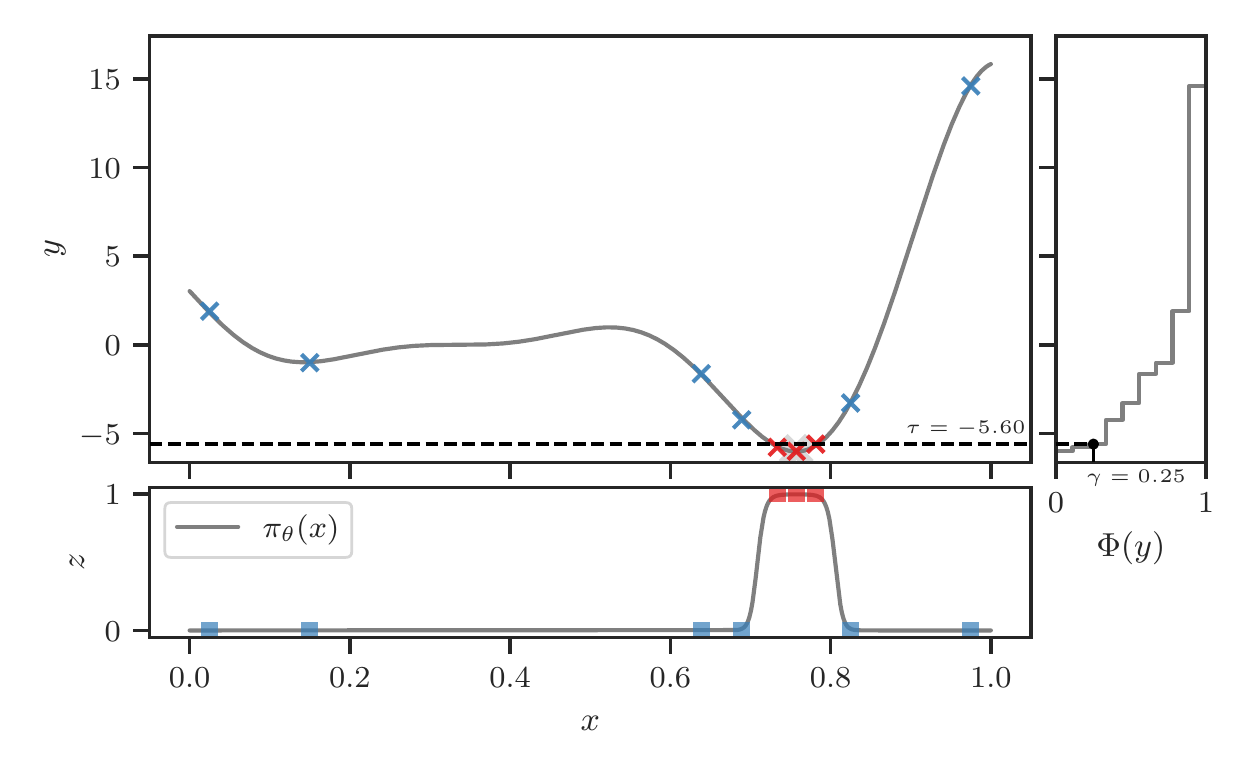}
    \caption{Iteration 6}
  \end{subfigure}
  \caption{We go through \cref{alg:bo-loop} step-by-step on a synthetic problem 
           for a half-dozen iterations. 
           Specifically, we minimize the \textsc{forrester} function.
           The algorithm is started 4 random initial designs. 
           Each subfigure depicts the state after \cref{line:parameters,line:inputs}---namely, 
           after \emph{updating} and \emph{maximizing} the classifier.
           In every subfigure, the main pane depicts the noise-free function, 
           represented by the \emph{solid gray} curve, and the set of 
           observations, represented by \emph{crosses} `$\times$'.
           The location that was evaluated in the previous iteration is 
           highlighted with a \emph{gray outline}.
           The right pane shows the \gls{ECDF} of the observed $y$ values. 
           The \emph{vertical dashed black line} in this pane is located at 
           $\gamma = \frac{1}{4}$. 
           The \emph{horizontal dashed black line} is located at $\tau$, the 
           value of $y$ such that $\Phi(y) = \frac{1}{4}$, i.e., 
           $\tau = \Phi^{-1} \left ( \frac{1}{4} \right )$.
           The instances below this horizontal line are assigned binary 
           label $z=1$, while those above are assigned $z=0$. 
           This is visualized in the bottom pane, alongside the probabilistic 
           classifier $\pi_\mbtheta(\mbx)$, 
           represented by the \emph{solid gray} curve.
           Finally, the maximizer of the classifier is represented by the 
           \emph{vertical solid green} line---this denotes the location to be 
           evaluated in the next iteration.}
  \label{fig:animation}
\end{figure}

For illustration purposes, we go through 
\cref{alg:bo-loop} step-by-step on a synthetic problem for a half-dozen 
iterations.
Specifically, we minimize the \textsc{forrester} function
\begin{equation*} 
  f(x) \defeq (6x-2)^2 \sin{(12x-4)},
\end{equation*}
in the domain $x \in [0, 1]$ with observation noise 
$\varepsilon \sim \cN(0, 0.05^2)$.
See \cref{fig:animation}.

\newpage

\newpage

\section{Properties of the classification problem}
\label{sec:properties_of_the_classification_problem}

We outline some notable properties of the \gls{BORE} classification 
problem as alluded to in~\cref{sec:methodology}.

\parhead{Class imbalance.}
By construction, this problem has class balance rate $\gamma$.

\begin{figure}[H]
  \centering
  \begin{subfigure}[t]{0.32\textwidth}
    \centering
    \includegraphics[width=\columnwidth]{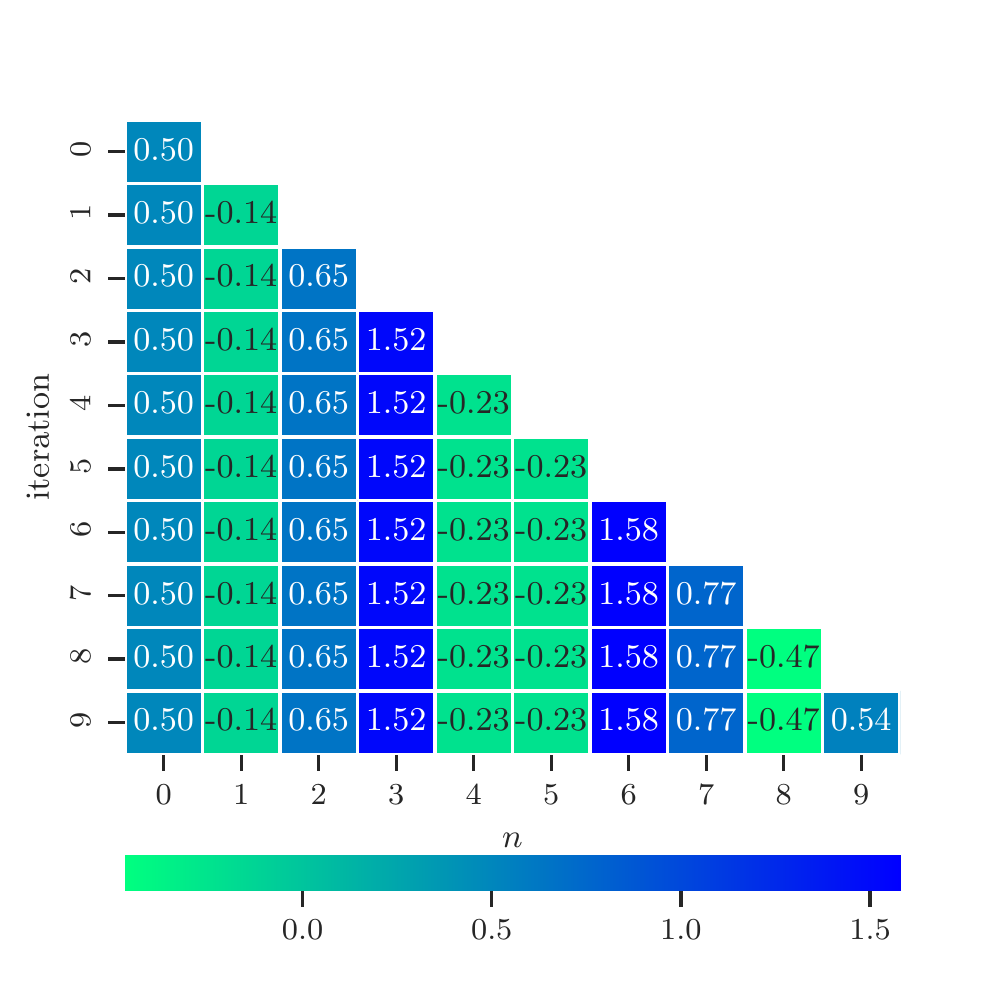}
    \caption{Continuous targets $y_n$}
  \end{subfigure}
  \begin{subfigure}[t]{0.32\textwidth}
    \centering
    \includegraphics[width=\columnwidth]{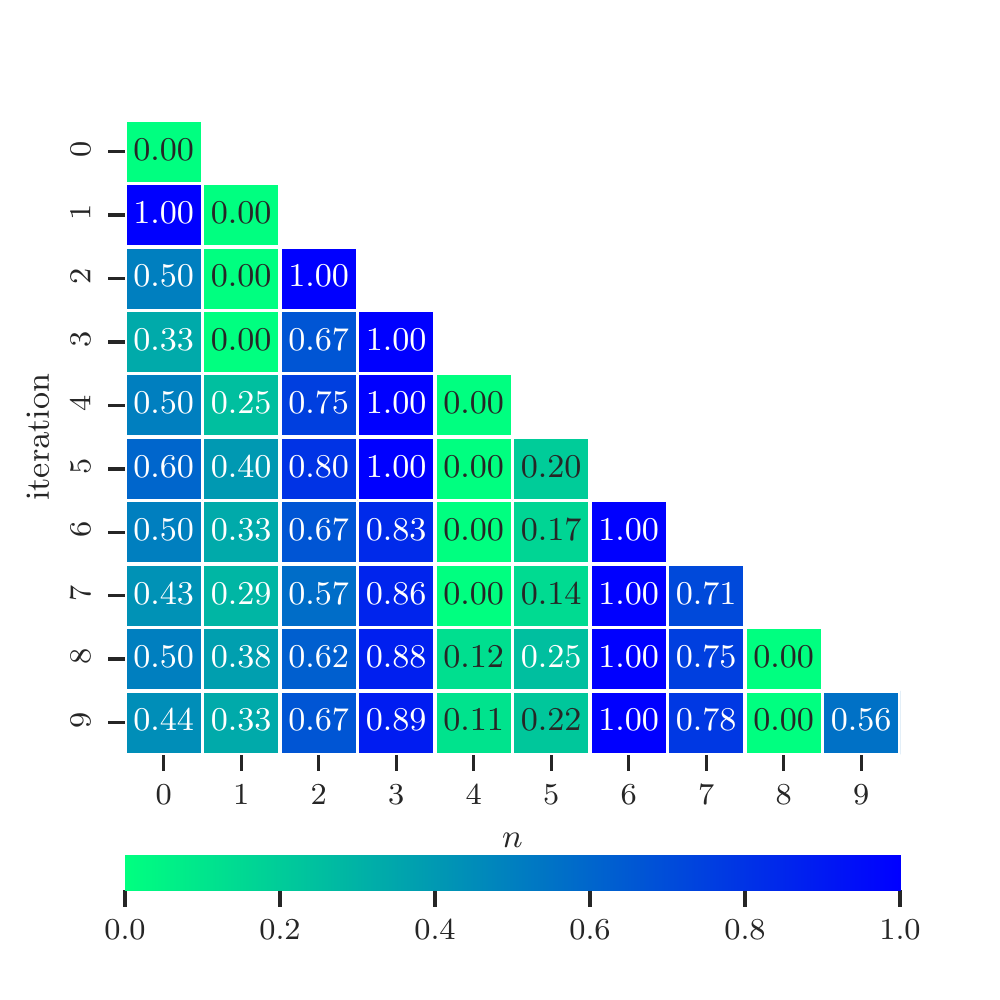}
    \caption{Empirical distribution $\Phi(y_n)$}
  \end{subfigure}
  \begin{subfigure}[t]{0.32\textwidth}
    \centering
    \includegraphics[width=\columnwidth]{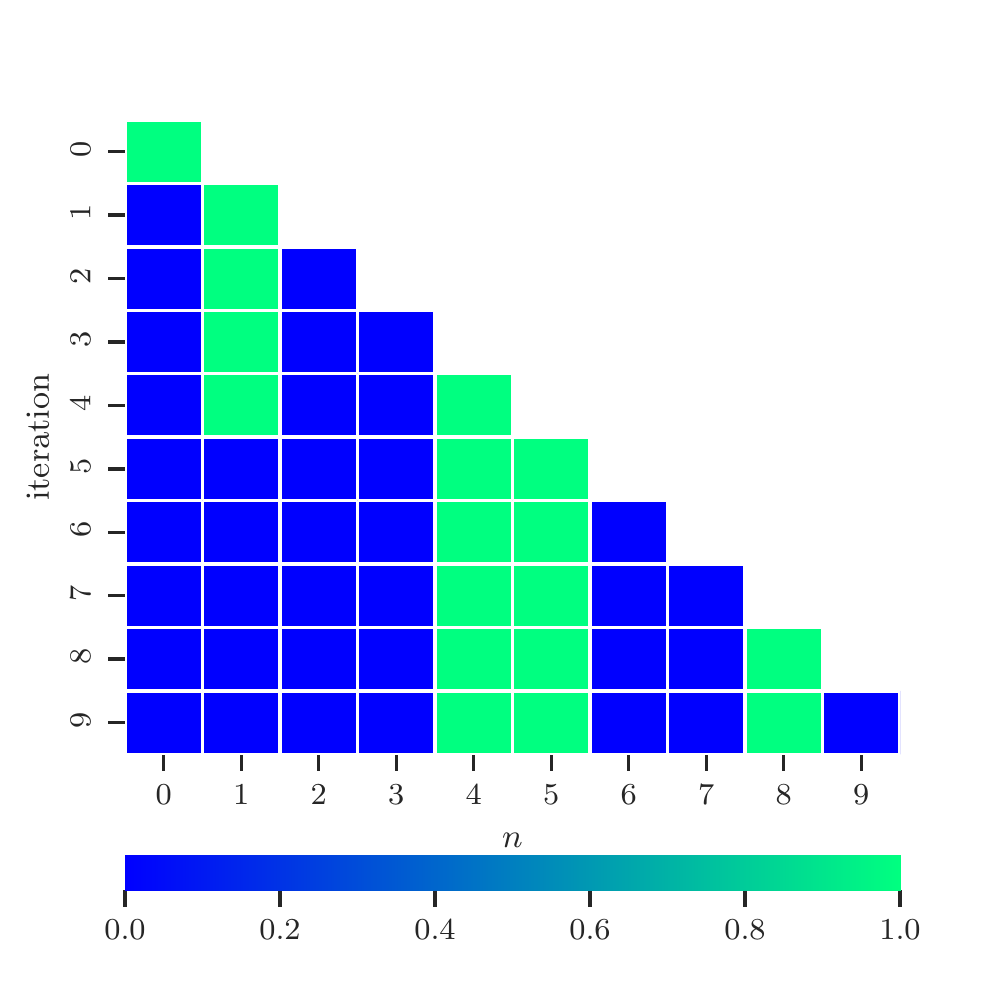}
    \caption{Binary labels $z_n$}
  \end{subfigure}
  \caption{Optimizing a ``noise-only'' synthetic function $f(x) = 0$
           with observation noise $\varepsilon \sim \cN(0, 1)$.
           The proportion is set to $\gamma=\nicefrac{1}{4}$.
           As we iterate through the \gls{BO} loop from top to bottom, 
           the array of targets grows from left to right. 
           Suffice it to say, in each iteration the size of the array increases 
           by one, resulting in a re-shuffling of the rankings and, by extension, quantiles.
           This in turn leads to
           the label for \emph{at most} one instance to flip. 
           Hence, between consecutive iterations, changes to the 
           classification dataset are fairly incremental.
           This property can be exploited to make classifier training more 
           efficient in each iteration.}
  \label{fig:label-evolution}
\end{figure}

\parhead{Label changes across iterations.} 
Assuming the proportion $\gamma$ is fixed across iterations, then,
in each iteration, 
we are guaranteed the following changes:
\begin{enumerate}
  \item a new input and its corresponding output $(\mbx_N, y_N)$ will be added
    to the dataset, thus
  \item creating a shift in the rankings and, by extension, quantiles of the 
    observed $y$ values, in turn
  \item leading to the binary label of \emph{at most} one instance to flip.
\end{enumerate}
Therefore, between consecutive iterations, changes to the classification dataset 
are fairly incremental.
This property can be exploited to make classifier training more efficient, 
especially in families of classifiers for which re-training entirely from scratch 
in each iteration may be superfluous and wasteful.
See \cref{fig:label-evolution} for an illustrated example.

Some viable strategies for reducing per-iteration classier learning overhead 
may include speeding up convergence by (i) \emph{importance sampling} (e.g. re-weighting 
new samples and those for which the label have flipped), (ii) \emph{early-stopping} 
(stop training early if either the loss or accuracy have not changed for some 
number of epochs) and (iii) \emph{annealing} (decaying the number of epochs or batch-wise 
training steps as optimization progresses).

\newpage

\section{Ablation studies}
\label{sec:ablation_studies}

\subsection{Maximizing the acquisition function}
\label{sub:maximizing_the_acquisition_function}

We examine different strategies for maximizing the acquisition function
(i.e. the classifier) in the tree-based variants of \gls{BORE}, 
namely, \acrshort{BORERF} and \acrshort{BOREXGB}. 
Decision trees are difficult to maximize since their response surfaces are 
discontinuous and nondifferentiable. 
Hence, we consider the following methods: \gls{RS} 
and \gls{DE}.
For each method, we further consider different evaluation budgets, i.e. limits 
on the number of evaluations of the acquisition function.
Specifically, we consider the limits $50, 100, 200,$ and $500$.

\begin{figure}[ht]
  \centering
  \begin{subfigure}[t]{0.48\textwidth}
    \centering
    \includegraphics[width=\linewidth]{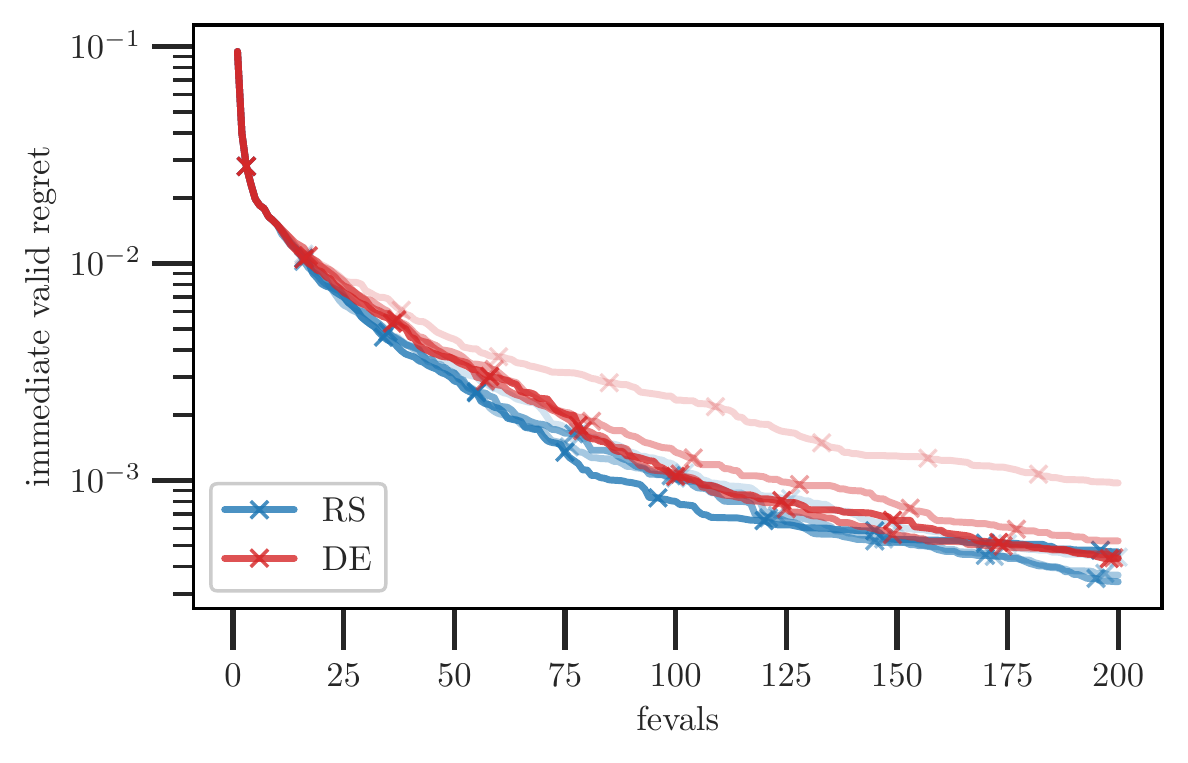}
    \caption{\acrshort{BORERF} on CIFAR-10}
    \label{fig:nasbench_ablation_rf}
  \end{subfigure}
  ~
   \begin{subfigure}[t]{0.48\textwidth}
    \centering
     \includegraphics[width=\linewidth]{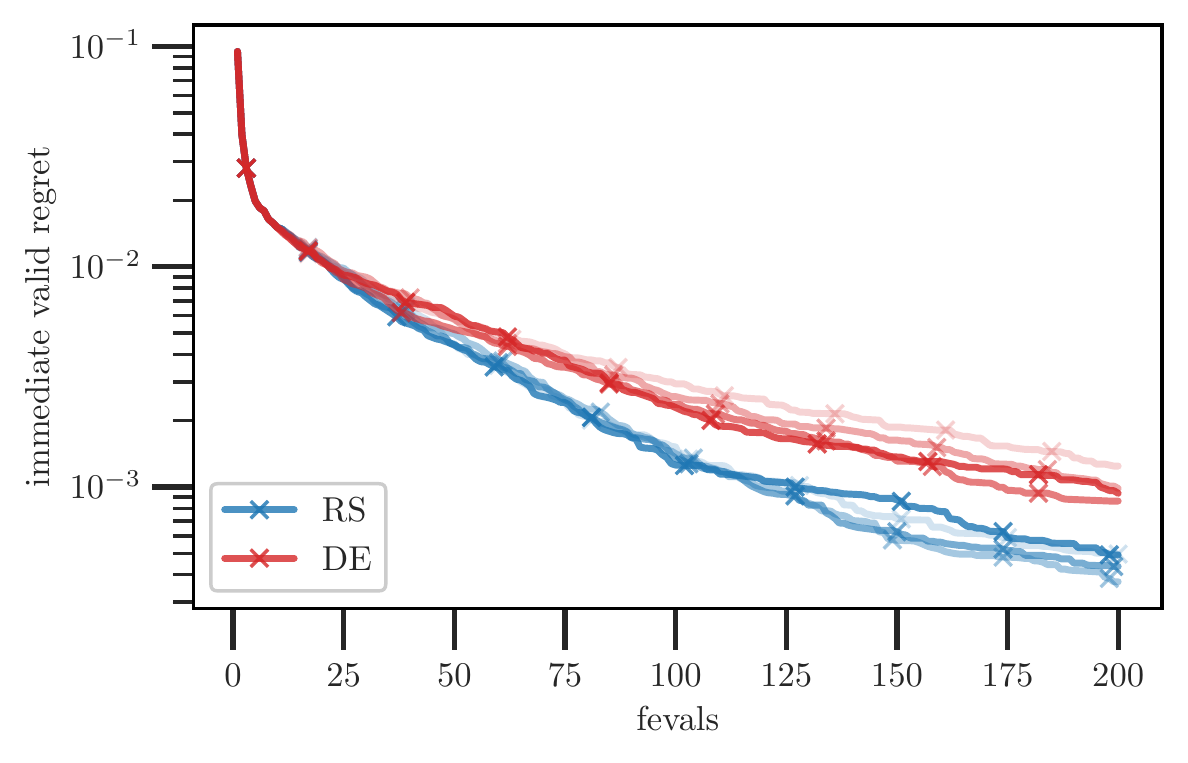}
     \caption{\acrshort{BOREXGB} on CIFAR-10}
     \label{fig:nasbench_ablation_xgboost}
  \end{subfigure}
   ~
  \caption{A comparison of various acquisition optimization strategies on the NASBench201 problem.}
  \label{fig:nasbench_ablation}
\end{figure}
In \cref{fig:nasbench_ablation_rf,fig:nasbench_ablation_xgboost}, we show the 
results of \acrshort{BORERF} and \acrshort{BOREXGB}, respectively, on the 
CIFAR-10 dataset of the NASBench201 benchmark, as described in \cref{sub:nasbench}.
Each curve represents the mean across 100 repeated runs. The opacity is 
proportional to the function evaluation limit, with the most transparent having 
the lowest limit and the most opaque having the highest limit.
We find that \gls{RS} appears to outperform \gls{DE} by a narrow margin.
Additionally, for \gls{DE}, a higher evaluation limit appears to be somewhat 
beneficial, while the opposite holds for \gls{RS}.

\subsection{Effects of calibration}
\label{sub:effects_of_calibration}

\begin{figure}[h]
  \centering
  \includegraphics[width=0.48\linewidth]{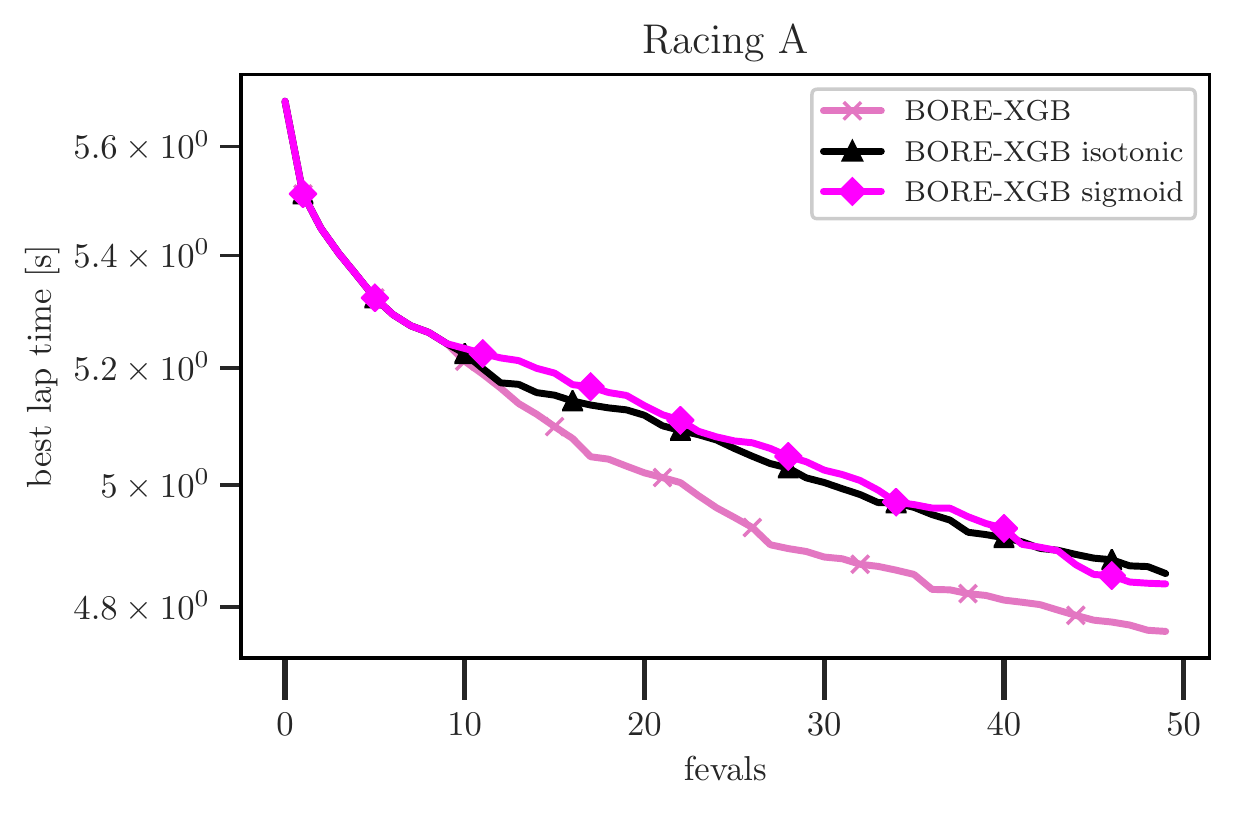}
  \caption{Effects of calibrating \gls{XGBOOST} in the \acrshort{BOREXGB} variant.
           Results of racing line optimization on the \textsc{uc berkeley} track.}
  \label{fig:calibration-racing-xgboost}
\end{figure}

\parhead{\acrshort{XGBOOST}.}
We apply the calibration approaches~\cite{niculescu2005predicting} we 
considered in \cref{sec:experiments} to \acrshort{XGBOOST} for the 
\acrshort{BOREXGB} variant, namely,
Platt scaling~\cite{platt1999probabilistic} and
isotonic regression~\cite{zadrozny2001obtaining,zadrozny2002transforming}.
As before, the results shown in \cref{fig:calibration-racing-xgboost} seem to 
suggest that these calibration methods have deleterious effects---at least when 
considering optimization problems which require only a small number of function
evaluations to reach the global minimum, since this yields a small dataset 
with which to calibrate the probabilistic classifier, making it susceptible to 
overfitting.


\newpage

\section{Toy example: relative density-ratio estimation by probabilistic classification}
\label{sec:toy_example}

\begin{figure}[H]
  \centering
  \begin{subfigure}[b]{0.48\textwidth}
      \centering
    \includegraphics[width=\columnwidth]{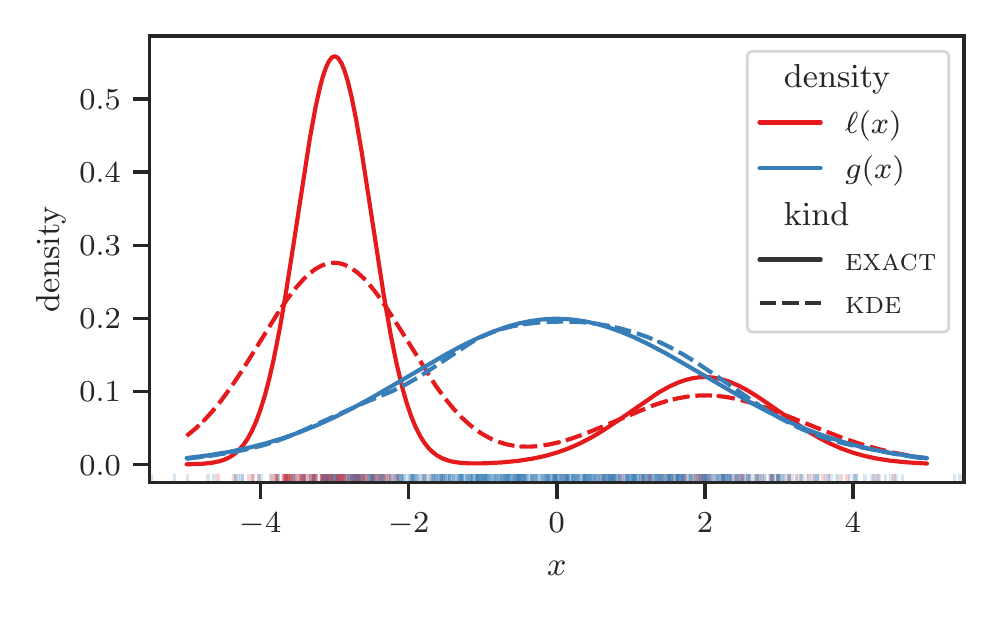}
    \caption{Densities $\ell(x)$ and $g(x)$.}
    \label{fig:toy-densities}
  \end{subfigure}%
  ~
  \begin{subfigure}[b]{0.48\textwidth} 
    \centering
    \includegraphics[width=\columnwidth]{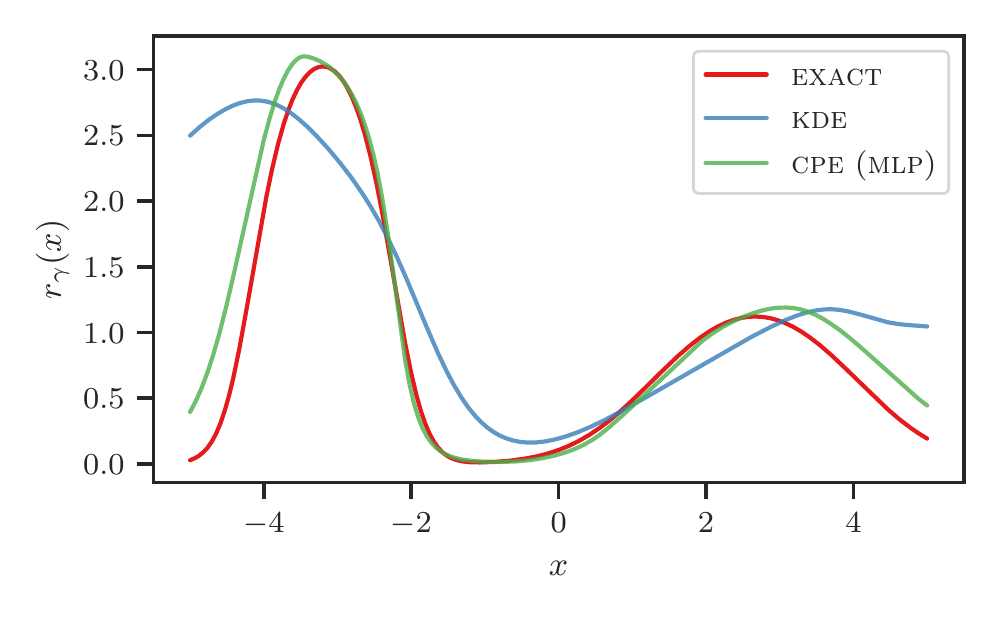}
    \caption{Relative density-ratio, estimated with an \acrshort{MLP} classifier.}
    \label{fig:toy-relative-density-ratios-mlp}
  \end{subfigure}
  \begin{subfigure}[b]{0.48\textwidth}
      \centering
    \includegraphics[width=\columnwidth]{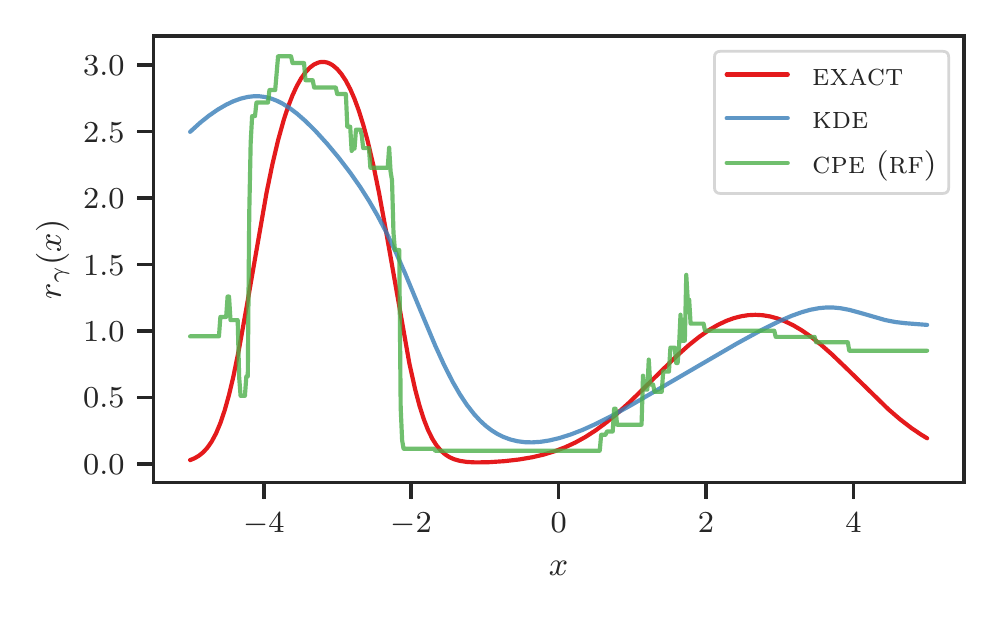}
    \caption{Relative density-ratio, estimated with a \acrshort{RF} classifier.}
    \label{fig:toy-relative-density-ratios-rf}
  \end{subfigure}%
  ~
  \begin{subfigure}[b]{0.48\textwidth} 
    \centering
    \includegraphics[width=\columnwidth]{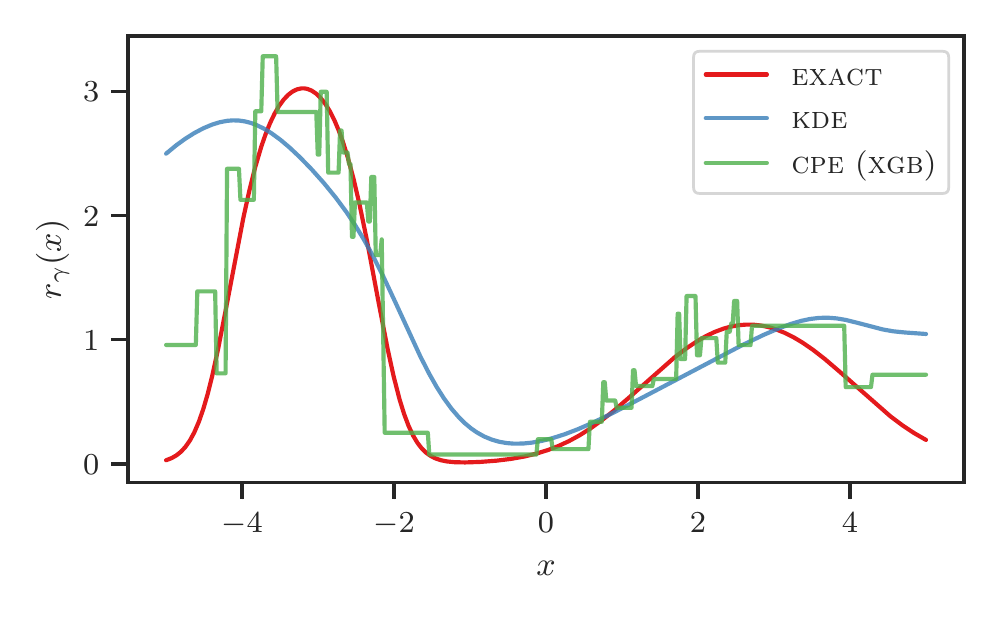}
    \caption{Relative density-ratio, estimated with an \acrshort{XGBOOST} classifier.}
    \label{fig:toy-relative-density-ratios-xgb}
  \end{subfigure}
  \caption{Synthetic toy example with (mixtures of) Gaussians.}
  \label{fig:toy}
\end{figure}

Consider the following toy example where the densities $\ell(x)$ and $g(x)$ are 
\emph{known} and given exactly by the following (mixture of) Gaussians,
\begin{equation*}
  \ell(x) \defeq 0.3 \cN(2, 1^2) + 0.7 \cN(-3, 0.5^2), 
  \qquad \text{and} \qquad 
  g(x) \defeq \cN(0, 2^2),
\end{equation*}
as illustrated by the \emph{solid red} and \emph{blue} lines in \cref{fig:toy-densities}, 
respectively.
We draw a total of $N=1000$ samples from these distributions, with a 
fraction $\gamma=\nicefrac{1}{4}$ drawn from $\ell(x)$ and the remainder 
from $g(x)$. 
These are represented by the vertical markers along the bottom of the $x$-axis
(a so-called ``rug plot'').
Then, two \glspl{KDE}, shown with \emph{dashed} lines, are fit on these 
respective sample sets, with kernel bandwidths selected according to the 
``normal reference'' rule-of-thumb\LT{citation needed?}. 
We see that, for both densities, the modes are recovered well, while for 
$\ell(x)$, the variances are overestimated in both of its mixture components.
As we shall see, this has deleterious effects on the resulting density ratio 
estimate.

In \cref{fig:toy-relative-density-ratios-mlp}, we represent the \emph{true} 
relative density-ratio with the \emph{red} line. 
The estimate resulting from taking the ratio of the \glspl{KDE} is shown in \emph{blue},
while that of the \gls{CPE} method described in \cref{sec:methodology} 
is shown in \emph{green}.
In this subfigure, the probabilistic classifier consists of a simple \gls{MLP}
with 3 hidden layers, each with and 32 units and $\mathtt{elu}$ activations.
In \cref{fig:toy-relative-density-ratios-rf,fig:toy-relative-density-ratios-xgb},
we show the same, but with \gls{RF} and \gls{XGBOOST} classifiers.

The \gls{CPE} methods appear, at least visually, to recover the exact density 
ratios well, whereas the \gls{KDE} method does so quite poorly. 
Perhaps the more important quality to focus on, for the purposes of \gls{BO}, 
is the \emph{mode} of the density-ratio functions. 
In the case of the \gls{KDE} method, we can see that this deviates 
significantly from that of the true density-ratio.
In this instance, even though \gls{KDE} fit $g(x)$ well and recovered the modes 
of $\ell(x)$ accurately, a slight overestimation of the variance in the latter 
led to a significant shift in the maximum of the resulting density-ratio 
functions.

\section{Implementation of Baselines}
\label{sec:baselines_implementations}

The software implementations of the baseline methods considered in our 
comparisons are described in \cref{tab:implementations}.
\begin{table}[ht]
\caption{Implementations of baseline methods.}
\label{tab:implementations}
\vskip 0.15in
\begin{center}
\begin{small}
\begin{tabular}{llll}
\toprule
Method          & Software Library      & URL (\texttt{github.com/*}) & Notes \\
\midrule
\acrshort{TPE}  & \textsf{HyperOpt}     & \texttt{\href{https://github.com/hyperopt/hyperopt}{hyperopt/hyperopt}}          & \\
\acrshort{SMAC} & \textsf{SMAC3}        & \texttt{\href{https://github.com/automl/SMAC3}{automl/SMAC3}}                    & \\
\textsc{gp-bo}  & \textsf{AutoGluon}    & \texttt{\href{https://github.com/awslabs/autogluon}{awslabs/autogluon}}          & in \texttt{autogluon.searcher.GPFIFOSearcher} \\
\acrshort{DE}   & \textsf{-}            & \texttt{-}                  & custom implementation \\
\acrshort{RE}   & \textsf{NASBench-101} & \texttt{\href{https://github.com/automl/nas_benchmarks}{automl/nas\_benchmarks}} & in \texttt{experiment\_scripts/run\_regularized\_evolution.py} \\
\bottomrule
\end{tabular}
\end{small}
\end{center}
\vskip -0.1in
\end{table}

\section{Experimental Set-up and Implementation Details}
\label{sec:details_implementations}

\parhead{Hardware.} 
In our experiments, we employ \texttt{m4.xlarge} \textsc{aws ec2} instances, 
which have the following specifications:
\begin{itemize}
  \item \textbf{CPU:} Intel(R) Xeon(R) E5-2676 v3 (4 Cores) @ 2.4 GHz
  \item \textbf{Memory:} 16GiB (DDR3)
\end{itemize}
\parhead{Software.}
Our method is implemented as a \emph{configuration generator} plug-in in the 
\href{https://github.com/automl/HpBandSter}{HpBandSter} library of 
\citet{falkner2018bohb}. 
The code will be released as open-source software upon publication.

The implementations of the classifiers on which the proposed variants of 
\gls{BORE} are based are described in \cref{tab:classifier-implementations}.
\begin{table}[ht]
\caption{Implementations of classifiers.}
\label{tab:classifier-implementations}
\vskip 0.15in
\begin{center}
\begin{small}
\begin{tabular}{lll}
\toprule
Model             & Software Library      & URL \\
\midrule
\Acrfull{MLP}     & \textsf{Keras}        & \url{https://keras.io} \\
\Acrfull{RF}      & \textsf{scikit-learn} & \url{https://scikit-learn.org} \\
\Acrfull{XGBOOST} & \textsf{XGBoost}        & \url{https://xgboost.readthedocs.io} \\
\bottomrule
\end{tabular}
\end{small}
\end{center}
\vskip -0.1in
\end{table}

We set out with the aim of devising a practical method that is not only 
agnostic to the  of choice classifier, but also robust to underlying
implementation details---down to the choice of algorithmic settings.
Ideally, any instantiation of \gls{BORE} should work well out-of-the-box 
without the need to tweak the sensible default settings that are typically 
provided by software libraries.
Therefore, unless otherwise stated, we emphasize that made no effort was made 
to adjust any settings and all reported results were obtained using the defaults.
For reproducibility, we explicitly enumerate them in turn for 
each of the proposed variants.

\subsection{\textsc{bore-rf}} 

We limit our description to the most salient hyperparameters. We do not deviate from 
the default settings which, at the time of this writing, are:
\begin{itemize}
  \item {\it number of trees} -- 100
  \item {\it minimum number of samples required to split an internal node} (\texttt{min\_samples\_split}) -- 2
  \item {\it maximum depth} -- unspecified (nodes are expanded until all leaves contain less than \texttt{min\_samples\_split} samples) 
\end{itemize}

\subsection{\textsc{bore-xgb}}

\begin{itemize}
  \item {\it number of trees (boosting rounds)} -- 100
  \item {\it learning rate} ($\eta$) -- $0.3$
  \item {\it minimum sum of instance weight (Hessian) needed in a child} (\texttt{min\_child\_weight}) -- 1
  \item {\it maximum depth} -- 6
\end{itemize}

\subsection{\textsc{bore-mlp}}
In the \acrshort{BOREMLP} variant, the classifier is a \gls{MLP} with 2 hidden 
layers, each with 32 units.
We consistently found $\mathtt{elu}$ activations \cite{clevert2015fast} to be 
particularly effective for lower-dimensional problems, with $\mathtt{relu}$ 
remaining otherwise the best choice.
We optimize the weights with \acrshort{ADAM} \citep{kingma2014adam} using 
batch size of $B=64$. 
For candidate suggestion, we optimize the input of the 
classifier 
wrt to its output using multi-started \acrshort{LBFGS} with three random 
restarts.

\begin{figure}[ht]
  \centering
  \includegraphics[width=\linewidth]{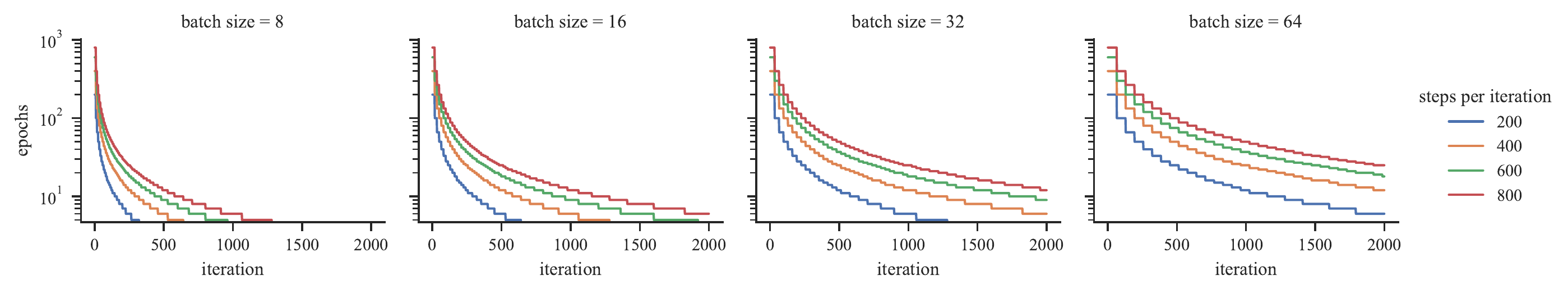}
  \caption{Effective number of epochs on the $n$th iteration for different 
  settings of batch size $B$ and number of steps per epoch $S$.}
  \label{fig:epoch_decay}
\end{figure}

\parhead{Epochs per iteration.}
To ensure the training time on \gls{BO} iteration $N$ is nonincreasing as a 
function of $N$, instead of directly specifying the number of epochs 
(i.e. full passes over the data), 
we specify the number of (batch-wise gradient) steps $S$ to train for in each 
iteration. 
Since the number of steps per epoch is $M = \lceil \nicefrac{N}{B} \rceil$, the
effective number of epochs on the $N$-th \gls{BO} iteration is then 
$E = \lfloor \nicefrac{S}{M} \rfloor$. 
For example, if $S=800$ and $B=64$, the number of epochs for iteration $N=512$ 
would be $E=100$. As another example, for all $0 < N \leq B$ (i.e. we have yet 
to observe enough data to fill a batch), we have $E=S=800$.
See \cref{fig:epoch_decay} for a plot of the effective number of epochs against 
iterations for different settings of batch size $B$ and number of steps per epoch 
$S$.
Across all our experiments, we fix $S=100$.

\section{Details of Benchmarks}
\label{sec:benchmarks_details}

\subsection{HPOBench}
\label{sub:hpobench}

The hyperparameters for the HPOBench problem and their ranges are summarized 
in \cref{tab:hpobench}.
\begin{table}[ht]
\caption{Configuration space for HPOBench.}
\label{tab:hpobench}
\vskip 0.15in
\begin{center}
\begin{small}
\begin{sc}
\begin{tabular}{lll}
\toprule
\multicolumn{2}{l}{Hyperparameter}         & Range \\
\midrule
\multicolumn{2}{l}{Initial \gls{LR}}       & $\{ \num{5e-4}, \num{1e-3}, \num{5e-3}, \num{1e-2}, \num{5e-2}, \num{1e-1} \}$ \\
\multicolumn{2}{l}{\gls{LR} schedule}      & $\{ \mathtt{cosine}, \mathtt{fixed} \}$ \\
\multicolumn{2}{l}{Batch size}             & $\{ 2^3, 2^4, 2^5, 2^6 \}$ \\
\multirow[t]{3}{*}{Layer 1} & Width        & $\{ 2^4, 2^5, 2^6, 2^7, 2^8, 2^9 \}$ \\
                            & Activation   & $\{ \mathtt{relu}, \mathtt{tanh} \}$ \\
                            & Dropout rate & $\{ 0.0, 0.3, 0.6 \}$ \\
\multirow[t]{3}{*}{Layer 2} & Width        & $\{ 2^4, 2^5, 2^6, 2^7, 2^8, 2^9 \}$ \\
                            & Activation   & $\{ \mathtt{relu}, \mathtt{tanh} \}$ \\
                            & Dropout rate & $\{ 0.0, 0.3, 0.6 \}$ \\
\bottomrule
\end{tabular}
\end{sc}
\end{small}
\end{center}
\vskip -0.1in
\end{table}
All hyperparameters are discrete---either ordered or unordered.
All told, there are $6 \times 2 \times 4 \times 6 \times 2 \times 3 \times 6 \times 2 \times 3 = 66,208$ possible combinations.
Further details on this problem can be found in \cite{klein2019tabular}. 

\subsection{NASBench201}
\label{sub:nasbench}

The hyperparameters for the HPOBench problem and their ranges are summarized 
in \cref{tab:nasbench201}.
\begin{table}[ht]
\caption{Configuration space for NASBench-201.}
\label{tab:nasbench201}
\vskip 0.15in
\begin{center}
\begin{small}
\begin{sc}
\begin{tabular}{ll}
\toprule
Hyperparameter & Range \\
\midrule
Arc 0 & $\{ \mathtt{none}, \mathtt{skip}\text{-}\mathtt{connect}, \mathtt{conv}\text{-}\mathtt{1\times1}, \mathtt{conv}\text{-}\mathtt{3\times3}, \mathtt{avg}\text{-}\mathtt{pool}\text{-}\mathtt{3\times3} \}$ \\
Arc 1 & $\{ \mathtt{none}, \mathtt{skip}\text{-}\mathtt{connect}, \mathtt{conv}\text{-}\mathtt{1\times1}, \mathtt{conv}\text{-}\mathtt{3\times3}, \mathtt{avg}\text{-}\mathtt{pool}\text{-}\mathtt{3\times3} \}$ \\
Arc 2 & $\{ \mathtt{none}, \mathtt{skip}\text{-}\mathtt{connect}, \mathtt{conv}\text{-}\mathtt{1\times1}, \mathtt{conv}\text{-}\mathtt{3\times3}, \mathtt{avg}\text{-}\mathtt{pool}\text{-}\mathtt{3\times3} \}$ \\
Arc 3 & $\{ \mathtt{none}, \mathtt{skip}\text{-}\mathtt{connect}, \mathtt{conv}\text{-}\mathtt{1\times1}, \mathtt{conv}\text{-}\mathtt{3\times3}, \mathtt{avg}\text{-}\mathtt{pool}\text{-}\mathtt{3\times3} \}$ \\
Arc 4 & $\{ \mathtt{none}, \mathtt{skip}\text{-}\mathtt{connect}, \mathtt{conv}\text{-}\mathtt{1\times1}, \mathtt{conv}\text{-}\mathtt{3\times3}, \mathtt{avg}\text{-}\mathtt{pool}\text{-}\mathtt{3\times3} \}$ \\
Arc 5 & $\{ \mathtt{none}, \mathtt{skip}\text{-}\mathtt{connect}, \mathtt{conv}\text{-}\mathtt{1\times1}, \mathtt{conv}\text{-}\mathtt{3\times3}, \mathtt{avg}\text{-}\mathtt{pool}\text{-}\mathtt{3\times3} \}$ \\
\bottomrule
\end{tabular}
\end{sc}
\end{small}
\end{center}
\vskip -0.1in
\end{table}
The operation associated with each of the $\binom{4}{2}=6$ arcs can belong to 
one of five categories.
Hence, there are $5^6=15,625$ possible combinations of hyperparameter configurations.
Further details on this problem can be found in \cite{dong2020bench}. 

\subsection{Robot pushing control}
\label{sub:robot_pushing_control}

This problem is concerned with tuning the controllers of two robot hands, 
with the goal of each pushing an object to some prescribed goal location
$\mbp_{g}^{(1)}$ and $\mbp_{g}^{(2)}$, respectively.
Let 
$\mbp_{s}^{(1)}$ and $\mbp_{s}^{(2)}$ denote the specified starting positions, and 
$\mbp_{f}^{(1)}$ and $\mbp_{f}^{(2)}$ the final positions (the latter of which are 
functions of
the control parameters $\mbx$).
The reward is defined as
\begin{equation*}
  R(\mbx) 
  \defeq 
  \underbrace{\| \mbp_{g}^{(1)} - \mbp_{s}^{(1)} \| + 
              \| \mbp_{g}^{(2)} - \mbp_{s}^{(2)} \|}_\text{initial distances}
  - \underbrace{(\| \mbp_{g}^{(1)} - \mbp_{f}^{(1)} \| + 
                 \| \mbp_{g}^{(2)} - \mbp_{f}^{(2)} \|)}_\text{final distances},
\end{equation*}
which effectively quantifies the amount of progress made toward pushing the 
objects to the desired
goal.
For each robot, the control parameters include the location and orientation 
of its hands, the pushing speed, moving direction and push duration.
These parameters and their ranges are summarized in \cref{tab:robot_pushing_control}.
\begin{table}[ht]
\caption{Configuration space for the robot pushing control problem.}
\label{tab:robot_pushing_control}
\vskip 0.15in
\begin{center}
\begin{small}
\begin{sc}
\begin{tabular}{lll}
\toprule
\multicolumn{2}{l}{Hyperparameter} & Range \\
\midrule
\multirow[t]{7}{*}{Robot 1} & Position $x$      & $[-5, 5]$ \\
                            & Position $y$      & $[-5, 5]$ \\
                            & Angle    $\theta$ & $[0, 2 \pi]$ \\
                            & Velocity $v_x$    & $[-10, 10]$ \\
                            & Velocity $v_y$    & $[-10, 10]$ \\
                            & Push Duration     & $[2, 30]$ \\
                            & Torque            & $[-5, 5]$ \\
\multirow[t]{7}{*}{Robot 2} & Position $x$      & $[-5, 5]$ \\
                            & Position $y$      & $[-5, 5]$ \\
                            & Angle    $\theta$ & $[0, 2 \pi]$ \\
                            & Velocity $v_x$    & $[-10, 10]$ \\
                            & Velocity $v_y$    & $[-10, 10]$ \\
                            & Push Duration     & $[2, 30]$ \\
                            & Torque            & $[-5, 5]$ \\
\bottomrule
\end{tabular}
\end{sc}
\end{small}
\end{center}
\vskip -0.1in
\end{table}

Further details on this problem can be found in~\cite{wang2018batched}.
This simulation is implemented with the \href{https://box2d.org/}{Box2D} library, and the 
associated code repository can be found at~\url{https://github.com/zi-w/Ensemble-Bayesian-Optimization}.

\subsection{Racing line optimization}
\label{sub:racing_line_optimization}

This problem is concerned with finding the optimal racing line. 
Namely, given a racetrack and a vehicle with known dynamics, 
the task is to determine a 
trajectory around the track for which the minimum time required to traverse it
is minimal.
We adopt the set-up of~\citet{jain2020computing}, who consider 
1:10 and 1:43 scale miniature remote-controlled cars traversing 
tracks at UC Berkeley~\cite{liniger2015optimization} and ETH Z\"{u}rich~\cite{rosolia2019learning},
respectively.

The trajectory is represented by a cubic spline parameterized by the 2D coordinates
of $D$ waypoints, each placed at locations along the length of the track, where 
the $i$th waypoint deviates from the centerline of the track by 
$x_i \in \left [- \frac{W}{2}, \frac{W}{2} \right ]$, for some track width $W$.
Hence, the parameters are the distances by which each waypoint deviates from the 
centerline, $\mbx = [ x_1 \cdots x_D ]^{\top}$.

Our blackbox function of interest, namely, the minimum time to traverse a given 
trajectory, is determined by the solution to a convex optimization problem 
involving \glspl{PDE}~\cite{lipp2014minimum}.
Further details on this problem can be found in~\cite{jain2020computing}, and the
associated code repository can be found at~\url{https://github.com/jainachin/bayesrace}.



\section{Parameters, hyperparameters, and meta-hyperparameters}

We explicitly identify the parameters $\mbomega$, hyperparameters $\mbtheta$, 
and meta-hyperparameters $\mblambda$ in our approach, making clear their distinction, 
examining their roles in comparison with other methods and discuss their treatment.

\glsreset{GP}
\glsreset{BNN}
\glsreset{NN}

\begin{table}[ht]
\caption{A taxonomy of parameters, hyperparameters, and meta-hyperparameters.}
\label{tab:hyperparameters}
\vskip 0.15in
\begin{center}
\begin{small}
\begin{tabular}{lp{1in}p{1in}p{1in}}
\toprule
& \acrshort{BO} with \glspl{GP} & \acrshort{BO} with \glspl{BNN} & \gls{BORE} with \glspl{NN} \\
\midrule
Meta-hyperparameters $\mblambda$ 
&
kernel family, 
kernel isotropy (\acrshort{ARD}),
etc.
& 
layer depth, widths, activations, etc.
& 
prior precision $\alpha$, likelihood precision $\beta$,
layer depth, widths, activations, etc.
\\
\midrule
Hyperparameters $\mbtheta$ 
&
kernel lengthscale and amplitude, $\ell$ and $\sigma$,
likelihood precision $\beta$ 
& 
prior precision $\alpha$, likelihood precision $\beta$ 
& 
weights $\mbW$, biases $\mbb$ 
\\
\midrule
Parameters $\mbomega$
& 
None $\varnothing$ (nonparametric)
& 
weights $\mbW$, biases $\mbb$ 
& 
None $\varnothing$ (by construction) \\
\bottomrule
\end{tabular}
\end{small}
\end{center}
\vskip -0.1in
\end{table}

\subsection{Parameters}
\label{sub:parameters}

\LT{The purpose of this section is to elaborate on related work in specific 
details, and to provide motivation for our work}

Since we seek to \emph{directly} approximate the acquisition function, 
our method is, by design, free of parameters $\mbomega$.
By contrast, in classical \gls{BO},
the acquisition function is derived from the analytical 
properties of the posterior predictive $p(y \g \mbx, \mbtheta, \cD_N)$.
To compute this, the uncertainty about parameters $\mbomega$ must be 
marginalized out 
\begin{equation} \label{eq:posterior-predictive}
  p(y \g \mbx, \mbtheta, \cD_N) 
  = \int p(y\g\mbx, \mbomega, \mbtheta) p(\mbomega \g \cD_N, \mbtheta) \, \mathrm{d}\mbomega,
  \quad
  \text{where}
  \quad
  p(\mbomega \g \cD_N, \mbtheta) 
  = \frac{p(\mby \g \mbX, \mbomega, \mbtheta) p(\mbomega \g \mbtheta)}{p(\mby \g \mbX, \mbtheta)}.
\end{equation}
While \glspl{GP} are free of parameters, the latent function values $\mbf$
must be marginalized out
\begin{equation*}
  p(y \g \mbx, \mbtheta, \cD_N) 
  = \int p(y\g\mbx, \mbf, \mbtheta) p(\mbf \g \cD_N, \mbtheta) \, \mathrm{d}\mbf.
\end{equation*}
In the case of \gls{GP} regression, this is easily computed by applying 
straightforward rules of Gaussian conditioning.
Unfortunately, few other models enjoy this luxury. 

\parhead{Case study: \glspl{BNN}.} As a concrete example, consider \glspl{BNN}. 
The parameters $\mbomega$ consist of the weights $\mbW$ and biases $\mbb$ in 
the \gls{NN}, while the hyperparameters $\mbtheta$ consist of the prior and 
likelihood precisions, $\alpha$ and $\beta$, respectively.
In general, $p(\mbomega \g \cD_N, \mbtheta)$ is not analytically tractable. 
\begin{itemize}
  \item To work around this, 
  \acrshort{DNGO}~\cite{snoek2015scalable} and 
  \acrshort{ABLR}~\cite{perrone2018scalable} both constrain the parameters 
  $\mbomega$ to include the weights and biases of only the \emph{final} 
  layer, $\mbW_{L}$ and $\mbb_{L}$, and relegate those of all preceding 
  layers, $\mbW_{1:L-1}$ and $\mbb_{1:L-1}$, to the hyperparameters $\mbtheta$. 
  This yields an exact (Gaussian) expression for $p(\mbomega \g \cD_N, \mbtheta)$ and
  $p(y \g \mbx, \mbtheta, \cD_N)$.
  To treat the hyperparameters,~\citet{perrone2018scalable} estimate $\mbW_{1:L-1}$, $\mbb_{1:L-1}$, $\alpha$ 
  and $\beta$ using type-II \gls{MLE}, while~\citet{snoek2015scalable} use a combination of type-II \gls{MLE} and 
  slice sampling~\cite{neal2003slice}.
  \item In contrast, \acrshort{BOHAMIANN}~\cite{springenberg2016bayesian} makes 
  no such simplifying distinctions regarding the layer weights and biases. 
  Consequently, they must resort to sampling-based approximations of
  $p(\mbomega \g \cD_N, \mbtheta)$, in their case by adopting \gls{SGHMC}~\cite{chen2014stochastic}.
\end{itemize}
In both approaches, compromises needed to be made in order to negotiate
the computation of $p(\mbomega \g \cD_N, \mbtheta)$. 
This is not to mention the problem of computing the posterior over 
hyperparameters $p(\mbtheta\g\cD_N)$, which we discuss next.
In contrast, \gls{BORE} avoids the problems associated with computing the 
posterior predictive $p(y \g \mbx, \mbtheta, \cD_N)$ and, by extension,
posterior $p(\mbomega \g \cD_N, \mbtheta)$ of \cref{eq:posterior-predictive}.
Therefore, such compromises are simply unnecessary.

\subsection{Hyperparameters}
\label{sub:hyperparameters}


For the sake of notational simplicity, we have thus far not been explicit 
about how the acquisition function depends on the hyperparameters $\mbtheta$
and how they are handled.
We first discuss generically how hyperparameters $\mbtheta$ are treated in \gls{BO}.
Refer to \cite{shahriari2015taking} for a full discussion.
In particular, we rewrite the \gls{EI} function, expressed in \cref{eq:expected-improvement-generic}, 
to explicitly include $\mbtheta$
\begin{equation*} 
  \alpha_{\gamma}(\mbx; \mbtheta, \cD_N) 
  \defeq \bbE_{p(y \g \mbx, \mbtheta, \cD_N)}[U(\mbx, y, \tau)].
\end{equation*}
\parhead{Marginal acquisition function.}
Ultimately, one wishes to maximize the \emph{marginal} acquisition 
function $A_{\gamma}(\mbx; \cD_N)$, which marginalizes out the uncertainty 
about the hyperparameters,
\begin{equation*} 
  A_{\gamma}(\mbx; \cD_N) 
  = \int \alpha_{\gamma}(\mbx; \cD_N, \mbtheta) p(\mbtheta\g\cD_N) \, \mathrm{d}\mbtheta
  \quad
  \text{where}
  \quad
  p(\mbtheta\g\cD_N)
  = \frac{p(\mby \g \mbX, \mbtheta) p(\mbtheta)}{p(\mby \g \mbX)}.
\end{equation*}
This consists of an expectation over the posterior $p(\mbtheta\g\cD_N)$ which 
is, generally speaking, analytically intractable. 
In practice, the most straightforward way to compute $A_{\gamma}(\mbx; \cD_N)$ 
is to approximate the posterior using a delta measure centered at some point 
estimate $\hat{\mbtheta}$,
either the type-II \gls{MLE} $\hat{\mbtheta}_{\textsc{mle}}$
or the \gls{MAP} estimate $\hat{\mbtheta}_{\textsc{map}}$.
This leads to
\begin{equation*}
  A_{\gamma}(\mbx; \cD_N) 
  \simeq \alpha_{\gamma}(\mbx; \cD_N, \hat{\mbtheta}).
\end{equation*}
Suffice it to say, sound uncertainty quantification is paramount to guiding 
exploration.
Since point estimates fail to capture uncertainty about hyperparameters $\mbtheta$,
it is often beneficial to turn instead to \gls{MC} estimation~\cite{snoek2012practical}
\begin{equation*}
  A_{\gamma}(\mbx; \cD_N) 
  \simeq
  \frac{1}{S} \sum_{s=1}^{S} \alpha_{\gamma}(\mbx; \cD_N, \mbtheta^{(s)}),
  \qquad \mbtheta^{(s)} \sim p(\mbtheta\g\cD_N).
\end{equation*}

\parhead{Marginal class-posterior probabilities.}
Recall that the likelihood of our model is
\begin{equation*}
p(z \g \mbx, \mbtheta) \defeq \Bernoulli(z \g \pi_\mbtheta(\mbx)),
\end{equation*}
or more succinctly $\pi_\mbtheta(\mbx) = p(z = 1 \g \mbx, \mbtheta)$.
We specify a prior $p(\mbtheta)$ on hyperparameters $\mbtheta$ and marginalize 
out its uncertainty to produce our analog to the marginal acquisition function
\begin{equation*}
    \Pi(\mbx; \cD_N) 
    = \int \pi_\mbtheta(\mbx) p(\mbtheta\g\cD_N) \, \mathrm{d}\mbtheta,
    \quad
    \text{where}
    \quad
    p(\mbtheta\g\cD_N)
    = \frac{p(\mbz \g \mbX, \mbtheta) p(\mbtheta)}{p(\mbz \g \mbX)}.
\end{equation*}
As in the generic case, we are ultimately interested in maximizing the 
marginal class-posterior probabilities $\Pi(\mbx; \cD_N)$.
However, much like $A_{\gamma}(\mbx; \cD_N)$, the marginal $\Pi(\mbx; \cD_N)$ 
is analytically intractable in turn due to the intractability of $p(\mbtheta\g\cD_N)$.
In this work, we focus on minimizing the log loss of \cref{eq:log-loss-empirical}, 
which is proportional to the negative log-likelihood
\begin{equation*}
\cL(\mbtheta)
= - \frac{1}{N} \sum_{n=1}^{N} \log p(z_n \g \mbx_n, \mbtheta) 
\propto - \log p(\mbz \g \mbX, \mbtheta).
\end{equation*}
Therefore, we're effectively performing the equivalent of type-II \gls{MLE}, 
\begin{equation*}
\hat{\mbtheta}_{\textsc{mle}} = \argmin_\mbtheta \cL(\mbtheta) = \argmax_\mbtheta \log p(\mbz \g \mbX, \mbtheta).
\end{equation*}
In the interest of improving exploration and, of particular importance in our 
case, calibration of class-membership probabilities, it may be beneficial to 
consider \gls{MC} and other approximate inference 
methods~\cite{blundell2015weight,gal2016dropout,lakshminarayanan2016simple}.
This remains fertile ground for future work.

\subsection{Meta-hyperparameters}
\label{sub:meta_hyperparameters}

In the case of \acrshort{BOREMLP}, the meta-hyperparameters might consist 
of, e.g. layer depth, widths, 
activations, etc---the tuning of which is often the reason one appeals to 
\gls{BO} in the first place.
For improvements in calibration, and therefore sample diversity, it may be 
beneficial to marginalize out the uncertainty about these, or considering some
approximation thereof, such as hyper-deep ensembles~\cite{wenzel2020hyperparameter}.



\end{document}